\documentclass{article}

\usepackage{iclr2027_conference,times}

\usepackage{amsmath,amsfonts,bm}

\def\eqref#1{equation~\ref{#1}}

\def\1{\bm{1}}

\def\rt{{\textnormal{t}}}

\DeclareMathAlphabet{\mathsfit}{\encodingdefault}{\sfdefault}{m}{sl}
\SetMathAlphabet{\mathsfit}{bold}{\encodingdefault}{\sfdefault}{bx}{n}

\usepackage{booktabs}
\usepackage{float}
\usepackage{capt-of}
\usepackage{fvextra}
\usepackage{graphicx}
\usepackage{tikz}
\usepackage{wrapfig,needspace}
\usepackage{microtype}
\usepackage{placeins}
\usepackage{xcolor}
\usepackage{hyperref}
\hypersetup{colorlinks=true,linkcolor=blue!50!black,citecolor=blue!50!black,urlcolor=blue!50!black}
\usepackage{url}
\usepackage{titletoc}

\newcommand{\bench}{\textsc{Gym-Anything-RT}}

\title{Gym-Anything-RT: Scaling Real-Time\\Environments for Vision-Language Agents}

\iclrfinalcopy
\author{Pranjal Aggarwal \& Sean Welleck \\
Carnegie Mellon University \\
\texttt{\{pranjala, swelleck\}@cs.cmu.edu}}

\begin{document}
\raggedbottom

\maketitle
\lhead{Preprint}

\begin{abstract}

AI agents have made substantial progress in domains where they have a non-trivial amount of time to deliberate between actions, such as mathematics, software engineering, and performing tasks on desktop applications. It is less clear how such models fare in settings that demand \textit{real-time interaction}, such as playing a video game or operating a physical device. In these settings, the environment state may evolve as the agent is reasoning, or even as the underlying model is performing a forward pass.
To gain insight into the real-time interaction capabilities of vision-language agents, we introduce \bench{}.
Our key idea is to construct \textit{visual minigames} that demand a portfolio of capabilities required for interactive real-time settings.
\bench{} uses an agentic pipeline to automatically generate such environments based on a taxonomy of six capabilities required for real-time interaction, including the ability to perceive the environment and act under an evolving environment state.
A key feature of \bench{} is that it allows for controlled comparison of real-time interaction capabilities: we can run the same minigame in real time, or pause or slow down the environment to relax the requirement for real-time actions.
We use \bench{} to construct 100 real-time minigames and evaluate AI agents on them. We find large capability gaps in real-time interaction. For example, GPT-6 Astra (xHigh) achieves a pass rate of only 9.0\% on the real-time minigames, which rises to 75.0\% when the environment is paused.
We also identify intriguing behaviors by the most powerful frontier agents: for instance, when given the ability to write code, these agents write code that bypasses the need to perceive and act in real time.
We publicly release \bench{} and the generated minigames to spur further research on AI agents in real-time interactive settings.\footnote{\raggedright Code and data are available at \url{https://github.com/gym-anything/gym-anything-rt}.}
\end{abstract}

\section{Introduction}

AI agents have made substantial progress in domains where they have time to deliberate between actions, such as mathematics~\citep{deepseekai2025deepseekr1}, software engineering~\citep{jimenez2024swebench}, and performing tasks on desktop applications~\citep{yuan2026osworld2,aggarwal2026gymanythingturnsoftwareagent,jang2026odysseys}. However, these agents have a key limitation: they are slow, often taking more than ten seconds to choose a single action. This is acceptable when the environment waits for the agent, but not in settings that demand \textit{real-time interaction}, such as playing a video game or operating a physical device~\citep{zhang2026videogamebenchvisionlanguagemodelscomplete,black2025realtimeexecutionactionchunking}. There, the environment keeps changing while the agent reasons, or even while the model performs a forward pass. By the time an action executes, it may no longer be valid, and the agent cannot react to what happened in the meantime. \looseness=-1

\begin{figure}[!t]
    \centering
    \includegraphics[width=\textwidth]{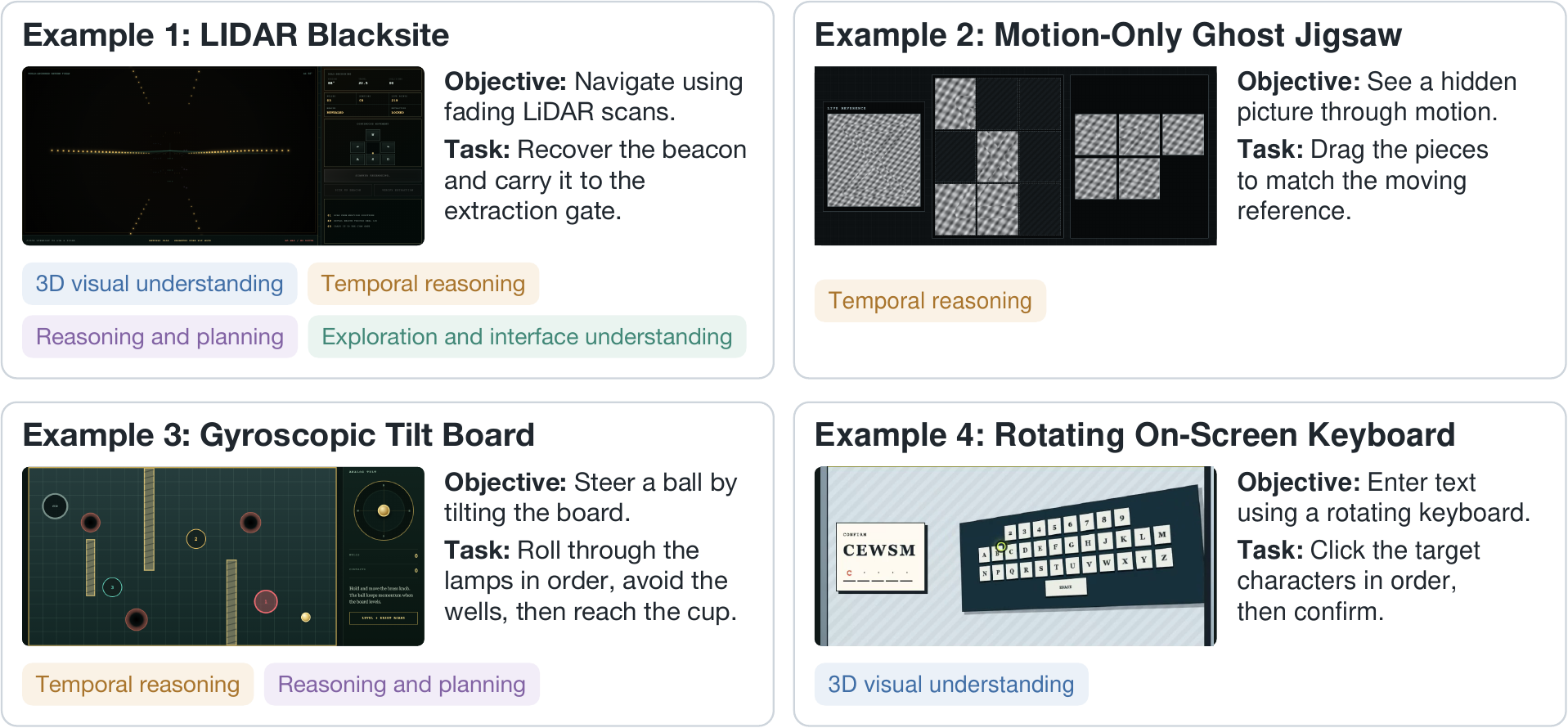}
    \caption{\textbf{Examples of real-time minigames generated with \bench{}.} The 100 minigames in our benchmark cover different combinations of capabilities, with controlled variations in difficulty and interaction.}
    \label{fig:teaser}
\end{figure}

Existing evaluations shed little light on this limitation. Computer-use benchmarks wait for the agent between actions, so they never test acting under an evolving state~\citep{xie2024osworldbenchmarkingmultimodalagents,yuan2026osworld2,aggarwal2026gymanythingturnsoftwareagent}. Video game benchmarks do run in real time. However, frontier models make little progress on them~\citep{zhang2026videogamebenchvisionlanguagemodelscomplete,ying2026aigamestorescalableopenended,ouyang2026gameworldstandardizedverifiableevaluation}. Furthermore, each game combines many capabilities, such as 3D spatial understanding, temporal reasoning, and learning unfamiliar controls, in a single fixed task.
This makes it difficult to tell whether models fail because they are too slow or because they lack some other capabilities. Separating these two causes requires controlled comparisons that vary how much time an agent has to respond on the same task.

To address these challenges, we introduce \bench{}, a framework for creating environments for studying the real-time interaction capabilities of vision-language agents. Our key idea is to construct \textit{visual minigames} that demand a portfolio of capabilities required for interactive real-time settings, while avoiding the complexity of complete games. For instance, in Rotating On-Screen Keyboard (Figure~\ref{fig:teaser}), the agent must enter a code by clicking keys that keep moving in three-dimensional space between its observation and its action. Furthermore, we build an evaluation stack that controls the clock of each minigame, so we can run the same task live at its natural speed, slow it down to give the agent more time to respond, or pause it between observation and action to remove the need for real-time actions. Since each minigame is a program, we can also modify it, for instance, to remove its temporal dynamics or to increase its difficulty. By comparing an agent's performance on these variants of the same task, we can tell whether it fails because it cannot act in time as the environment changes, or because the task is hard for other reasons.

Building such minigames by hand is laborious, since each needs its own visuals, dynamics, controls, and verifier. We therefore build an agentic pipeline that generates the minigames automatically. We first survey prior work on embodied agents~\citep{savva2019habitatplatformembodiedai,srivastava2021behaviorbenchmarkeverydayhousehold,yang2025embodiedbenchcomprehensivebenchmarkingmultimodal}, video games~\citep{zhang2026videogamebenchvisionlanguagemodelscomplete}, and computer use~\citep{cheng-etal-2024-seeclick,xie2024osworldbenchmarkingmultimodalagents} to identify six capabilities required for real-time interaction, such as 3D visual understanding, temporal reasoning, and acting under an evolving environment state, and select minigames that cover the capabilities. We then build on the creator-auditor procedure of Gym-Anything~\citep{aggarwal2026gymanythingturnsoftwareagent}, in which one coding agent builds an environment and an independent agent audits it. In our setting, a creator agent implements each minigame as a program, and additionally writes an oracle solution that uses the program's internal state to show that the task is solvable. Independent audit agents then test the minigame and report problems to the creator agent, which fixes them until the minigame behaves as intended. Finally, review agents label the capabilities that each minigame requires. Each program also defines four easier difficulty levels and a simplified interaction mode, which let us evaluate weaker models and study what makes a task hard.

Using this pipeline, we construct a benchmark of 100 real-time minigames and their non-real-time counterparts, and use it to evaluate frontier and open-weight models. We find that current agents perform poorly in real time. GPT-6 Astra (xHigh) solves only 9.0\% of the real-time tasks, while it solves 66.0\% of their non-real-time counterparts, which we obtain by modifying each program so that it no longer requires real-time interaction. Other models perform even worse: Claude Opus 5.5 (xHigh) and GPT-6 Luna (xHigh) solve only 5.0\% and 1.0\% of the real-time tasks, and the open-weight Qwen3.8-27B and Qwen3.5-9B solve only 1.0\% and 0.0\%, respectively. In addition, we find that for frontier models, the primary bottleneck is speed. On the same real-time tasks, slowing the environment by 64$\times$ raises Astra's pass rate from 9.0\% to 57.0\%, and pausing it raises the pass rate to 75.0\%. Since we also control difficulty and the interaction interface, we further identify temporal reasoning and exploration of unfamiliar interfaces as additional sources of difficulty. \looseness=-1

Interestingly, we also find that \textit{frontier agents can partially offset their poor real-time capabilities through their strong coding and reasoning capabilities}. Specifically, when we allow Astra to run code, it writes scripts that perceive and act on its behalf between its own responses. For instance, in Rotating On-Screen Keyboard, it locates the moving keys using image template matching and clicks them within 200\,ms. As a result, its real-time pass rate rises from 9.0\% to 31.0\%.

In summary, our contributions are three-fold. First, we introduce \bench{}, a framework that uses coding agents to automatically generate real-time visual minigames as programs, with oracle solutions and controlled difficulty and interaction. Its evaluation stack runs each minigame live, slowed down, or paused. Second, using \bench{}, we construct a benchmark of 100 real-time minigames and their non-real-time counterparts for studying the real-time interaction capabilities of vision-language agents. Third, through extensive evaluations, we show that current agents perform poorly in real time, that speed is the primary bottleneck for frontier models, and that frontier agents can partially offset their slow responses by writing code.

\section{Methodology}
\label{sec:methodology}

Our goal is to develop a suite of minigames that test the real-time interaction capabilities of vision-language agents in a controllable way. In particular, we construct minigames in which the time available to respond can be controlled while the underlying task stays the same. Formally, each minigame is an executable environment $\mathcal{E}$ whose program defines its internal state, controls, and clock. At each step, the agent observes a screenshot $o_t$ and issues mouse and keyboard actions $a_t$, while the environment keeps evolving as the agent chooses $a_t$. We say that a task requires \textit{real-time interaction} when the agent must use recent visual evidence to choose an action for the state in which it will execute, so that changes during the delay between observation and action can reduce that action's value (Appendix~\ref{app:real-time-definition}).

\begin{figure}[t]
    \centering
    \resizebox{\linewidth}{!}{%
    \begin{tikzpicture}[
        font=\small,
        stage/.style={draw, rounded corners=1pt, align=center,
            text width=2.1cm, minimum height=1.2cm, inner sep=4pt},
        flow/.style={-stealth, thick}
    ]
        \node[stage] (survey) at (0,0) {\textbf{Survey}\\Source records};
        \node[stage] (selection) at (2.7,0) {\textbf{Selection}\\Seed minigame $P$};
        \node[stage] (creator) at (5.4,0) {\textbf{Creator}\\Program $\mathcal{E}$\\Oracle and evidence};
        \node[stage] (auditor) at (8.1,0) {\textbf{Auditor}\\Verification};
        \node[stage] (review) at (10.8,0) {\textbf{Review}\\Difficulty and capabilities};
        \node[stage] (output) at (10.8,-2.0) {\textbf{Configurations}\\$\mathcal{E}_{d,i}$};
        \draw[flow] (survey) -- (selection);
        \draw[flow] (selection) -- (creator);
        \draw[flow] (creator) -- (auditor);
        \draw[flow] (auditor) -- (review);
        \draw[flow] (review) -- (output);
        \draw[flow] (auditor.south) -- (8.1,-1.25)
            -- node[below] {Revision feedback} (5.4,-1.25)
            -- (creator.south);
    \end{tikzpicture}%
    }
    \caption{\textbf{Overview of our construction method.} A creator agent implements selected minigames with controllable difficulty and interaction, constructs oracle solutions, and records evidence of their behavior. An independent auditor checks the implementation and evidence, returning issues for revision until the minigame passes the audit. Review agents calibrate difficulty and annotate capability requirements for controlled model evaluation.}
    \label{fig:construction-pipeline}
\end{figure}
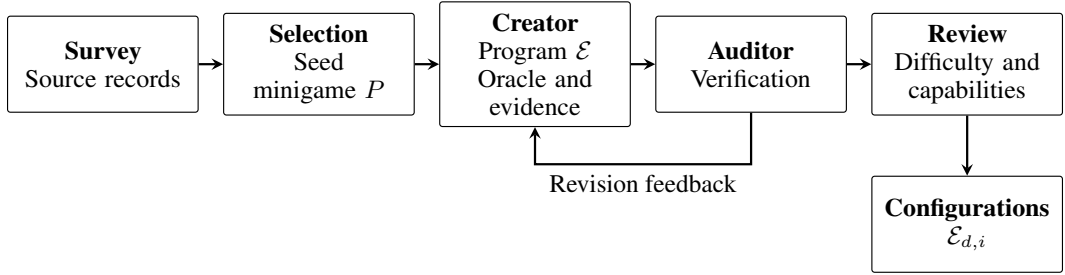

Each minigame has ten configurations $\mathcal{E}_{d,i}$, where $d \in \{1,\ldots,5\}$ specifies the difficulty and $i \in \{\mathrm{simplified},\mathrm{full}\}$ specifies the interaction mode. Each configuration can be run with a slowdown $s$: live execution at its natural speed ($s=1$), slowed execution ($s>1$), in which the clock advances at $1/s$ of its normal rate, and paused execution, in which the environment freezes until the agent acts. In addition, each real-time minigame has a non-real-time counterpart $\bar{\mathcal{E}}$, obtained through a minimal change to its timing (Section~\ref{sec:counterparts}). Since a broken or unsolvable minigame would be counted as an agent failure, every configuration must also be solvable and objectively gradable. We construct these minigames with coding agents, that is, LLM agents that can write and run code and inspect screenshots in a sandbox. All agents are instances of GPT-6 Astra run through the Codex CLI. Each agent takes on one of four roles: survey agents collect candidate minigames, creator agents implement real-time versions of them, audit agents verify them, and review agents calibrate difficulty and annotate capabilities. Figure~\ref{fig:construction-pipeline} summarizes this procedure.

\subsection{Task Discovery and Selection}
\label{sec:task-discovery}

We prompt survey agents to search prior work, public demonstrations, visual puzzles, online videos, and social media for challenges that satisfy four criteria: they can be formulated as self-contained minigames, require visual understanding and grounding, require actions through a visual interface, and have objectively verifiable outcomes. These include tasks that involve moving objects, hidden information, physical manipulation, or changes that occur between actions. For every source, the survey agent records its provenance, objective, visual structure, required interactions, temporal behavior, failure and recovery process, and information needed to verify success.

We manually review the first 15 minigames and use them as quality standards for subsequent agent iterations. We select candidates with a traceable source, distinct visual and interaction requirements, an objectively gradable outcome, and meaningful mouse or keyboard actions. A source need not involve real-time interaction itself, since the creator agent develops its mechanic into a real-time task (Section~\ref{sec:creator-auditor}). We exclude tasks dominated by OCR, static image classification, arithmetic, or question answering, since these tasks do not require the forms of visual interaction that we seek to measure.

\subsection{Creator-Auditor Construction}
\label{sec:creator-auditor}

To construct a real-time minigame $\mathcal{E}$ from a source $P$, we build upon the creator-auditor procedure from Gym-Anything~\citep{aggarwal2026gymanythingturnsoftwareagent}, in which a creator agent builds each environment and independent audit agents verify it. In our setting, the creator agent writes a program that implements the minigame and exposes a common interface for screenshot observations as well as mouse and keyboard actions. Crucially, we ask the creator agent to turn the task posed by the source into one that requires real-time interaction, as defined above, whether or not the source itself does. Adding decorative motion, a countdown, or faster animation is not sufficient. For example, Rotating On-Screen Keyboard requires the agent to click keys that keep moving between observation and action.

The creator agent implements the generator, interface, dynamics, action handling, and grader of $\mathcal{E}$. It also constructs an oracle solver that uses the privileged state of the environment to determine a valid solution and executes it through the same action space available to the agent. A successful oracle execution therefore demonstrates that the task is solvable and that its action and grading paths agree. The creator agent records the implementation, oracle trajectory, and visual and temporal evidence for these claims (Appendix~\ref{app:creator-audit}).

Each implementation is then audited by a fresh audit agent that cannot modify the program. The audit agent inspects the code and evidence, runs the environment when necessary, and checks generation, solvability, rendering, action handling, grading, and recovery from failure. It also checks the real-time requirement itself, searching for shortcuts that would let an agent avoid acting in real time, such as a brake that halts the moving parts or periodic motion that permits a precomputed solution. The audit agent returns issues to the creator agent for revision, and the loop repeats until an audit passes.

\subsection{Non-Real-Time Counterparts}
\label{sec:counterparts}

A low pass rate on a real-time minigame does not by itself show that real-time interaction is the cause, since the underlying task may also be hard. We therefore construct, for each real-time minigame $\mathcal{E}$, a non-real-time counterpart $\bar{\mathcal{E}}$ that poses the same task without requiring real-time interaction. The creator agent makes a minimal change to the timing of $\mathcal{E}$, while preserving its goal, controls, difficulty, and the information available to the agent. An independent audit agent then checks across all ten configurations that the two versions differ only in their timing. Comparing an agent on $\mathcal{E}$ and $\bar{\mathcal{E}}$ thus lets us separate the effect of the real-time requirement from the difficulty of the underlying task.

\subsection{Controlled Task Variations}
\label{sec:controlled-variations}

To also evaluate weaker models and study what makes a task hard, we construct easier variants of each minigame. We vary the difficulty $d$ and the interaction mode $i$ while holding the other and the random seed fixed, for both $\mathcal{E}$ and its counterpart.

\begin{figure}[!t]
    \centering
    \begin{minipage}[t]{0.32\textwidth}
        \centering
        \includegraphics[width=\linewidth]{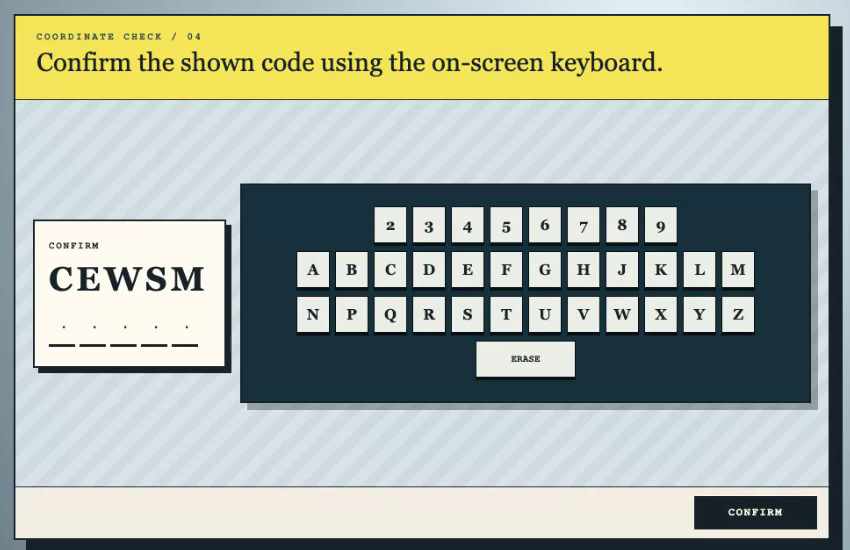}
        \vspace{-1mm}

        {\small\textbf{(a)} The code and keyboard before the first action.}
    \end{minipage}
    \hfill
    \begin{minipage}[t]{0.32\textwidth}
        \centering
        \includegraphics[width=\linewidth]{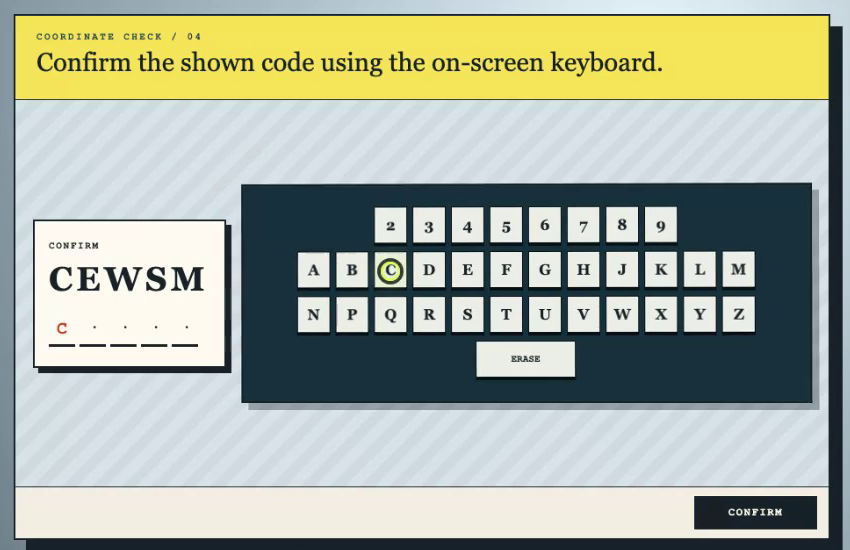}
        \vspace{-1mm}

        {\small\textbf{(b)} The first key is selected.}
    \end{minipage}
    \hfill
    \begin{minipage}[t]{0.32\textwidth}
        \centering
        \includegraphics[width=\linewidth]{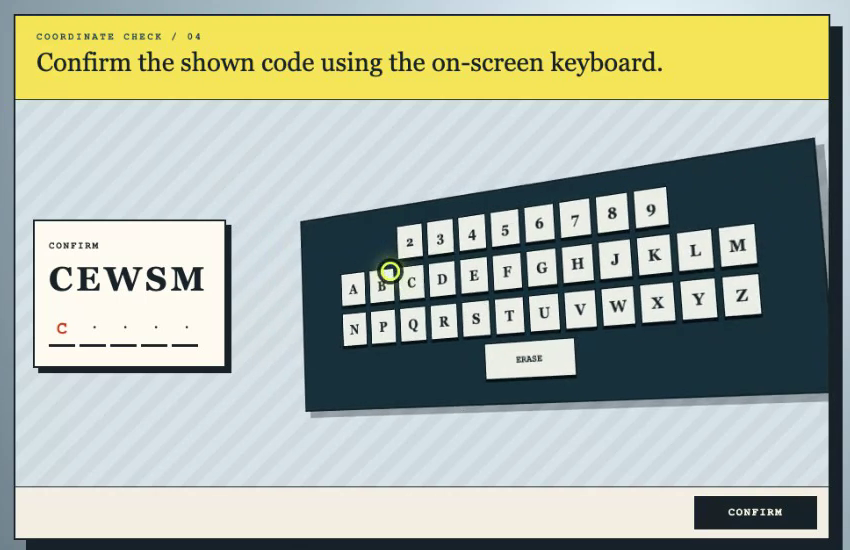}
        \vspace{-1mm}

        {\small\textbf{(c)} The keyboard moves before the next click.}
    \end{minipage}
    \caption{\textbf{Rotating On-Screen Keyboard.} The model must enter a code by clicking keys that move between observation and action.}
    \label{fig:rotating-keyboard}
\end{figure}

A difficulty-review agent first reads the generator, visible interface, grader, verifier, and oracle solver to determine what the original minigame requires. It compares this configuration with an approved set of reference minigames and assigns it to one of five levels. The creator agent preserves the original configuration at that level and constructs the remaining levels around it by changing the amount of information, number of dependencies, speed, precision, dynamics, or simultaneous control required for success. A difficulty-review agent then checks each adjacent pair and rejects any spurious difficulty increase, such as requesting repeated actions or adding ambiguity. The agent also ensures that the same types of perception, memory, decisions, and control are required across difficulty levels. For example, the levels of Rotating On-Screen Keyboard (Figure~\ref{fig:rotating-keyboard}) vary the code length, the number and size of keys, and the keyboard motion.

The creator agent also implements simplified and full interaction for every difficulty level. Simplified interaction exposes proxy controls, while full interaction requires the direct mouse or keyboard manipulation appropriate to the minigame. For example, in Rotating On-Screen Keyboard, simplified interaction lets the agent type the code directly with keyboard actions, while full interaction requires clicking the rotating keys. For a fixed seed and difficulty, the audit agent verifies that both modes generate the same minigame and differ only in the controls used to act on it. Appendices~\ref{app:difficulty} and~\ref{app:interaction} give examples of both variations.

\subsection{Controlling Time}
\label{sec:real-time-interaction}

To control how much time an agent has to respond, we build an evaluation stack on top of Gym-Anything~\citep{aggarwal2026gymanythingturnsoftwareagent} that controls the clock of each minigame. During live execution, the task continues to change from observation through action execution at its natural speed. In Rotating On-Screen Keyboard, for example, a key can move between observation and action, so the agent must account for its position when the click executes. Since any delay between an action and its effect counts against the agent, we also build a direct connection protocol. A central server still launches and manages many sandboxes in parallel, but each agent exchanges actions and screenshots over a direct connection to its own sandbox, bypassing the server. This keeps the median time from sending an action to receiving the resulting screenshot below 50\,ms in our timing tests (Appendix~\ref{app:infrastructure}). For slowed execution, the stack scales every clock of the minigame, including its timers, animations, and server-side clocks, so that a slowdown of $s$ advances the game by one second during $s$ seconds of real time. Action delivery and screenshot capture still run in real time, so the slowdown changes the time available to respond while the agent's interaction stays the same. During paused execution, the task advances through an observation window and then freezes on the final frame until the action is executed. All settings use the same minigame, seed, difficulty, interaction mode, and verifier. Appendix~\ref{app:real-time-protocol} describes the evaluation protocol.

\subsection{Capability Review}
\label{sec:capability-review}

To relate agent failures to the capabilities a minigame requires, a capability-review agent labels each configuration of both the real-time minigames and their counterparts for 3D visual understanding, temporal reasoning, reasoning and planning, and exploration and interface understanding (see Appendix~\ref{app:capability-criteria} for definitions). It reviews the game's program, interface, verifier, oracle solution, and recorded behavior to identify what a solution using screenshots and mouse or keyboard actions requires. Each label is supported by code or behavioral evidence. A real-time review agent checks the real-time requirement separately for each configuration, using the implementation and recorded behavior (Appendix~\ref{app:real-time-definition}).

\section{\bench{}}
\label{sec:benchmark}

We use the pipeline from Section~\ref{sec:methodology} to construct \bench{}, comprising 100 real-time minigames and their 100 non-real-time counterparts. Each also has four easier difficulty levels and a simplified interaction mode, giving 2,000 configurations in total. We evaluate models using the live, slowed, and paused settings defined in Section~\ref{sec:real-time-interaction}.

\subsection{Capability Coverage}
\label{sec:benchmark-capabilities}

\begin{wraptable}{r}{0.40\textwidth}
    \vspace{-12pt}
    \centering
    \caption{\textbf{Benchmark statistics.}}
    \label{tab:benchmark-stats}
    \small
    \begin{tabular}{lr}
        \toprule
        \textbf{Statistic} & \textbf{Count} \\
        \midrule
        Real-time minigames & 100 \\
        Non-real-time counterparts & 100 \\
        Difficulty levels & 5 \\
        Interaction modes & 2 \\
        Configurations & 2,000 \\
        \quad Real-time & 1,000 \\
        \quad Non-real-time & 1,000 \\
        \bottomrule
    \end{tabular}
    \vspace{-8pt}
\end{wraptable}

Table~\ref{tab:benchmark-stats} summarizes \bench{}. Using the labels from the capability review (Section~\ref{sec:capability-review}), 31\% of the real-time minigames require 3D visual understanding at some difficulty level, all of them require temporal reasoning, 96\% require reasoning and planning, and 52\% require exploration and interface understanding. Their counterparts require the same capabilities at similar rates, except temporal reasoning, which drops to 82\% once the real-time requirement is removed. Table~\ref{tab:minigame-list} lists all minigames and their capability labels.

\subsection{Evaluation Protocol}
\label{sec:benchmark-protocol}

We evaluate models by pass rate, time to solve, and inference cost, and compare these metrics across real-time minigames and their counterparts, slowdowns, difficulty levels, interaction modes, and capability requirements. Models receive task instructions and interact through screenshots, mouse actions, and keyboard input, without access to program internals or oracle solutions. Task-specific verifiers determine whether each attempt succeeds. Appendix~\ref{app:evaluation} describes the models, prompts, budgets, and cost computation.

\section{Results and Analysis}
\label{sec:results}
\providecommand{\resultstar}{\ensuremath{^{*}}}

\subsection{Experimental Setup}

We evaluate Qwen3.5-9B, Qwen3.8-27B, GPT-6 Luna with xHigh reasoning effort, Claude Opus 5.5 with xHigh effort, and GPT-6 Astra with low and xHigh reasoning effort. We evaluate models in the live setting with full interaction at the highest difficulty, and Section~\ref{sec:difficulty-results} reports results on the easier difficulty levels. Following prior work on computer-use agents \citep{xie2024osworldbenchmarkingmultimodalagents,aggarwal2025programmingpixelscomputeruseagents}, models observe the environment only through screenshots and act only through mouse and keyboard actions. Models cannot run code or access the page source or program internals. In the coding experiment, we additionally allow models to run code in an isolated sandbox, where they can request screenshots and execute the same mouse and keyboard actions through an API. Interaction with the environment remains the same. Further experiments vary difficulty, interaction mode, the real-time setting, and whether interface rules are provided.

\subsection{Current Agents Perform Poorly in Real Time}
\label{sec:main-results}

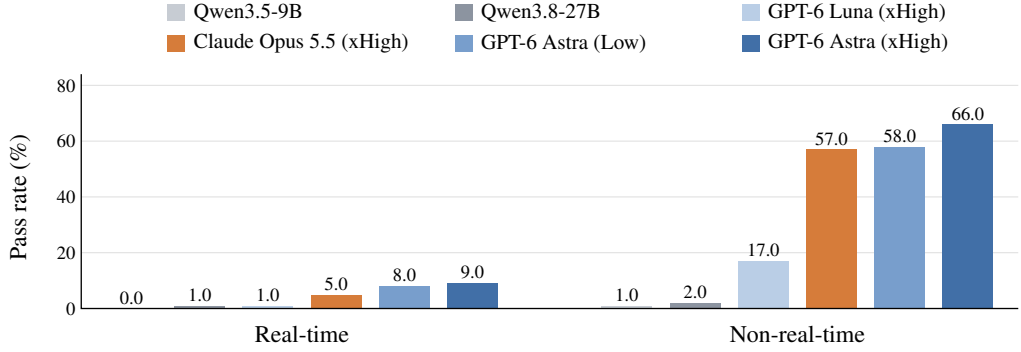
\begin{figure}[t]
    \centering
    \begin{tikzpicture}[x=0.068\linewidth,y=1.05pt,
        every node/.style={font=\fontsize{7}{8}\selectfont},
        qwensmall/.style={fill={rgb,255:red,199;green,204;blue,211}},
        qwenlarge/.style={fill={rgb,255:red,140;green,148;blue,160}},
        luna/.style={fill={rgb,255:red,186;green,206;blue,230}},
        opus/.style={fill={rgb,255:red,214;green,124;blue,58}},
        astralow/.style={fill={rgb,255:red,120;green,158;blue,204}},
        astraxhigh/.style={fill={rgb,255:red,65;green,111;blue,167}}]
        \foreach \tick in {0,20,40,60,80} {
            \draw[black!13,line width=0.35pt] (0,\tick) -- (13.05,\tick);
            \node[anchor=east,inner sep=2pt] at (0,\tick) {\tick};
        }
        \foreach \i/\s/\rt/\nrt in {0/qwensmall/0.0/1.0,1/qwenlarge/1.0/2.0,2/luna/1.0/17.0,3/opus/5.0/57.0,4/astralow/8.0/58.0,5/astraxhigh/9.0/66.0} {
            \fill[\s] (0.7+\i*0.95-0.35,0) rectangle (0.7+\i*0.95+0.35,\rt);
            \node[anchor=south,inner sep=1.5pt] at (0.7+\i*0.95,\rt) {\rt};
            \fill[\s] (7.6+\i*0.95-0.35,0) rectangle (7.6+\i*0.95+0.35,\nrt);
            \node[anchor=south,inner sep=1.5pt] at (7.6+\i*0.95,\nrt) {\nrt};
        }
        \draw[line width=0.5pt] (0,84) -- (0,0) -- (13.05,0);
        \node[rotate=90,font=\fontsize{9}{10}\selectfont] at (-0.85,40) {Pass rate (\%)};
        \node[font=\fontsize{9}{10}\selectfont] at (3.075,-9) {Real-time};
        \node[font=\fontsize{9}{10}\selectfont] at (9.975,-9) {Non-real-time};
        \foreach \x/\y/\s/\name in {1.2/106/qwensmall/Qwen3.5-9B,5.2/106/qwenlarge/Qwen3.8-27B,9.2/106/luna/GPT-6 Luna (xHigh),1.2/95/opus/Claude Opus 5.5 (xHigh),5.2/95/astralow/GPT-6 Astra (Low),9.2/95/astraxhigh/GPT-6 Astra (xHigh)} {
            \fill[\s] (\x,\y-3) rectangle (\x+0.25,\y+3);
            \node[anchor=west,inner sep=3pt,font=\fontsize{8}{9}\selectfont] at (\x+0.25,\y) {\name};
        }
    \end{tikzpicture}
    \caption{\textbf{Pass rates (\%) on \bench{} at difficulty 5 with full interaction in the live setting.} Each model is evaluated on the 100 real-time tasks and their 100 non-real-time counterparts. Even the strongest model performs poorly on tasks requiring real-time interaction.}
    \label{fig:main-results}
\end{figure}

Figure~\ref{fig:main-results} compares the pass rates of each model on the real-time tasks and on their non-real-time counterparts (Section~\ref{sec:real-time-interaction}). Even the strongest model, GPT-6 Astra (xHigh), solves only 9.0\% of the real-time tasks at the highest difficulty, compared with 66.0\% of their non-real-time counterparts. Interestingly, higher reasoning effort helps only when the task waits for the model: from Low to xHigh effort, Astra's pass rate rises from 58.0\% to 66.0\% on the non-real-time counterparts, but only from 8.0\% to 9.0\% on the real-time tasks. Claude Opus 5.5 (xHigh) solves 5.0\% of the real-time tasks and 57.0\% of their counterparts. GPT-6 Luna (xHigh) performs worse on both, solving 1.0\% of the real-time tasks and 17.0\% of their counterparts, and the open-weight Qwen3.8-27B and Qwen3.5-9B solve only 1.0\% and 0.0\% of the real-time tasks. These two models also solve only 2.0\% and 1.0\% of the counterparts, so their failures are not specific to real time. Since each real-time task and its counterpart share the same minigame, these results show that acting in real time, rather than the underlying task, accounts for most failures of frontier models. In the following sections, we examine why.

\subsection{Speed Is the Primary Bottleneck for Frontier Models}
\label{sec:speed-scaling}

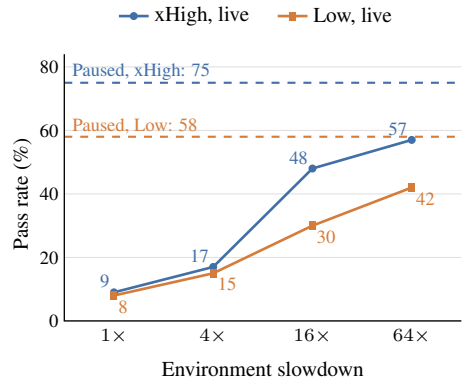
\begin{wrapfigure}{r}{0.47\textwidth}
    \vspace{-14pt}
    \centering
    \begin{tikzpicture}[x=0.2\linewidth,y=0.042cm,
        every node/.style={font=\fontsize{7}{8}\selectfont},
        xhigh/.style={color={rgb,255:red,65;green,111;blue,167}},
        low/.style={color={rgb,255:red,214;green,124;blue,58}}]
        \foreach \tick in {0,20,40,60,80} {
            \draw[black!13,line width=0.35pt] (0,\tick) -- (4,\tick);
            \node[anchor=east,inner sep=2pt] at (0,\tick) {\tick};
        }
        \draw[xhigh,dashed,line width=0.8pt] (0,75) -- (4,75);
        \draw[low,dashed,line width=0.8pt] (0,58) -- (4,58);
        \node[xhigh,anchor=south west,inner sep=1pt] at (0.05,75) {Paused, xHigh: 75};
        \node[low,anchor=south west,inner sep=1pt] at (0.05,58) {Paused, Low: 58};
        \draw[xhigh,line width=1pt] (0.5,9) -- (1.5,17) -- (2.5,48) -- (3.5,57);
        \draw[low,line width=1pt] (0.5,8) -- (1.5,15) -- (2.5,30) -- (3.5,42);
        \foreach \x/\v in {0.5/9,1.5/17,2.5/48,3.5/57} {
            \fill[xhigh] (\x,\v) circle[radius=1.5pt];
            \node[xhigh,anchor=south east,inner sep=1.5pt] at (\x,\v) {\v};
        }
        \foreach \x/\v in {0.5/8,1.5/15,2.5/30,3.5/42} {
            \fill[low] (\x-0.035,\v-1.4) rectangle (\x+0.035,\v+1.4);
            \node[low,anchor=north west,inner sep=1.5pt] at (\x,\v) {\v};
        }
        \draw[line width=0.5pt] (0,84) -- (0,0) -- (4,0);
        \foreach \x/\label in {0.5/$1\times$,1.5/$4\times$,2.5/$16\times$,3.5/$64\times$} {
            \node[anchor=north,inner sep=3pt] at (\x,0) {\label};
        }
        \node[font=\fontsize{8}{9}\selectfont] at (2,-15) {Environment slowdown};
        \node[rotate=90,font=\fontsize{8}{9}\selectfont] at (-0.42,40) {Pass rate (\%)};
        \draw[xhigh,line width=1pt] (0.55,96) -- (0.85,96);
        \fill[xhigh] (0.7,96) circle[radius=1.5pt];
        \node[anchor=west,inner sep=2pt,font=\fontsize{8}{9}\selectfont] at (0.85,96) {xHigh, live};
        \draw[low,line width=1pt] (2.15,96) -- (2.45,96);
        \fill[low] (2.265,94.6) rectangle (2.335,97.4);
        \node[anchor=west,inner sep=2pt,font=\fontsize{8}{9}\selectfont] at (2.45,96) {Low, live};
    \end{tikzpicture}
    \vspace{-4pt}
    \caption{\textbf{Effect of environment speed on Astra.} Pass rates increase as the environment slows, with xHigh remaining below its paused performance at $64\times$ slowdown.}
    \label{fig:speed-scaling}
    \vspace{-10pt}
\end{wrapfigure}
To study how response speed limits performance, we use our evaluation stack to slow down the environment while leaving the model's inference unchanged (Section~\ref{sec:real-time-interaction}). A slowdown of $s$ advances the environment clock at $1/s$ of its normal rate, giving the model more time to respond before the task changes, while pausing freezes the environment until the model acts. Figure~\ref{fig:speed-scaling} reports the pass rate of GPT-6 Astra on the same 100 real-time tasks at the highest difficulty as we slow the environment. With xHigh reasoning effort, pass rate rises steadily from 9\% at normal speed to 17\% at $4\times$, 48\% at $16\times$, and 57\% at $64\times$ slowdown, and reaches 75\% when the environment is paused. Pausing similarly raises GPT-6 Luna's pass rate on real-time tasks from 6.0\% to 29.8\% across all difficulty levels. Since the tasks are unchanged and only the time available to respond grows, these results show that for frontier models, slow responses are the primary reason for failure in real time.

Moreover, the benefit of additional time grows with reasoning effort. At normal speed, Low and xHigh effort perform similarly, 8\% and 9\%, since additional reasoning also takes additional time. As the environment slows, xHigh pulls ahead, reaching 57\% at $64\times$ compared with 42\% for Low, and 75\% compared with 58\% when paused. Interestingly, even at $64\times$ slowdown, Astra (xHigh) remains 18 points below its paused performance, suggesting that some tasks remain difficult as long as the environment keeps changing at all.

\subsection{Strong Coding Capabilities Partially Offset Weak Real-Time Capabilities}
\label{sec:coding-results}

\begin{table}[t]
    \centering
    \begin{minipage}[t]{0.51\textwidth}
        \centering
        \caption{\textbf{Effect of coding on real-time performance.} Astra benefits most from coding on tasks requiring real-time interaction.}
        \label{tab:coding-results}
        \vspace{4pt}
        \small
        \setlength{\tabcolsep}{3pt}
        \resizebox{\linewidth}{!}{%
        \begin{tabular}{@{}lrrrr@{}}
        \toprule
        & \multicolumn{2}{c}{\textbf{Real-time}} & \multicolumn{2}{c}{\textbf{Non-real-time}} \\
        \cmidrule(lr){2-3}\cmidrule(lr){4-5}
        \textbf{Model} & \textbf{Coding} & \textbf{Non-coding} & \textbf{Coding} & \textbf{Non-coding} \\
        \midrule
        GPT-6 Astra (Low) & 14.0 & 8.0 & 63.0 & 58.0 \\
        GPT-6 Astra (xHigh) & 31.0 & 9.0 & 74.0 & 66.0 \\
        GPT-6 Luna (xHigh) & 5.0 & 1.0 & 18.0 & 17.0 \\
        \bottomrule
    \end{tabular}}
    \end{minipage}\hfill
    \begin{minipage}[t]{0.46\textwidth}
        \centering
        \caption{\textbf{Efficiency on the real-time configurations at difficulty 5.} All models are slower and less successful than the human author, who is evaluated on 20 randomly sampled tasks.}
        \label{tab:efficiency-results}
        \vspace{4pt}
        \small
        \setlength{\tabcolsep}{3pt}
        \resizebox{\linewidth}{!}{%
        \begin{tabular}{@{}lrrr@{}}
        \toprule
        \textbf{Agent} & \textbf{Pass rate $\uparrow$} & \textbf{Avg. time $\downarrow$} & \textbf{Avg. cost $\downarrow$} \\
        & \textbf{(\%)} & \textbf{(min)} & \textbf{(\$)} \\
        \midrule
        Qwen3.5-9B & 0.0 & 75.9 & 0.62 \\
        Qwen3.8-27B & 1.0 & 63.0 & 0.64 \\
        GPT-6 Luna & 1.0 & 6.4 & 0.07 \\
        Claude Opus 5.5 & 5.0 & 15.8 & 6.40 \\
        GPT-6 Astra (Low) & 8.0 & 4.3 & 2.76 \\
        GPT-6 Astra (xHigh) & 9.0 & 7.2 & 4.50 \\
        Human & 45.0 & 2.2 & -- \\
        \bottomrule
    \end{tabular}}
    \end{minipage}
\end{table}

Since frontier models are strong coding agents, we next examine whether they can use code to compensate for their slow responses. We allow models to run code in an isolated sandbox, where they can request screenshots and execute the same mouse and keyboard actions through an API (Appendix~\ref{app:coding-protocol}). The action space is unchanged, but models can now write scripts that observe and act without waiting for the model's next response. Table~\ref{tab:coding-results} reports results at the highest difficulty with full interaction in the live setting. Coding raises the pass rate of Astra (xHigh) on real-time tasks from 9.0\% to 31.0\%, a 3.4-fold increase, compared with a gain of only 8.0 points on their non-real-time counterparts, from 66.0\% to 74.0\%. The gain also grows with model capability: Astra (Low) improves from 8.0\% to 14.0\%, and Luna only from 1.0\% to 5.0\%.

To understand this gain, we inspect Astra's trajectories. Astra writes image-processing scripts that perceive and act between its own responses. For instance, in Rotating On-Screen Keyboard, it locates the moving keys using image template matching and clicks them within 200\,ms. In Clockwork Clutch Safe, where moving gears must be aligned, it tracks gear markers and times clutch releases using estimates of their motion (Appendix~\ref{app:coding-examples}). In effect, the model moves the real-time part of the task into code, and reasons only about what the code should do. This suggests that similar scripts could help slow models act in other real-time domains, such as video games and physical interfaces. Nonetheless, coding still leaves Astra far below its paused performance of 75.0\%, and such scripts are difficult to write when the relevant visual information cannot be extracted reliably or prepared for in advance.

\subsection{Models Are Slow and Expensive}
\label{sec:efficiency-results}

Real-time applications require agents that are not only accurate, but also fast and affordable. Table~\ref{tab:efficiency-results} reports the pass rate, average time, and average inference cost per attempt on the 100 real-time tasks at the highest difficulty. Astra (xHigh) achieves the highest pass rate, 9.0\%, but takes 7.2 minutes and costs \$4.50 per attempt on average. Lowering its reasoning effort to Low reduces the time to 4.3 minutes and the cost to \$2.76, with a similar pass rate of 8.0\%, consistent with additional reasoning helping little when the environment does not wait. Claude Opus 5.5 (xHigh) costs the most, \$6.40 per attempt, and takes 15.8 minutes, while solving only 5.0\% of the tasks. Luna is far cheaper at \$0.07 per attempt, but takes 6.4 minutes and solves only 1.0\% of the tasks. Qwen3.8-27B and Qwen3.5-9B cost \$0.64 and \$0.62 per attempt, but take about ten times longer than Luna, 63.0 and 75.9 minutes, while solving only 1.0\% and 0.0\% of the tasks. As a case study, we compare these results with those of a human author who, like the evaluated models, sees each minigame for the first time. On 20 randomly sampled tasks, the human solves 9 tasks, or 45.0\%, in an average of 2.2 minutes per task (Appendix~\ref{app:human}). This is five times the pass rate of Astra (xHigh), in less than a third of its average time. These results show that progress in real-time interaction requires improving not only the pass rate of models, but also their speed and cost~\citep{abhyankar2026osworldhumanbenchmarkingefficiencycomputeruse,black2025realtimeexecutionactionchunking}.

\subsection{Pausing Also Helps When Temporal Reasoning Is Required}
\label{sec:temporal-results}

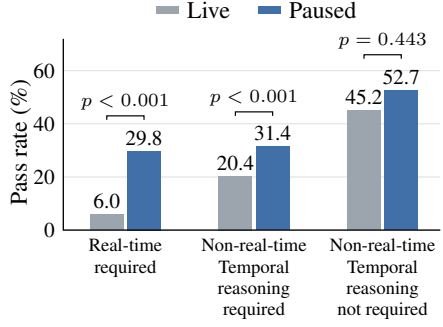
\begin{figure}[t]
    \centering
    \begin{minipage}[c]{0.48\textwidth}
    \centering
    \begin{tikzpicture}[x=0.175\linewidth,y=10pt,
        every node/.style={font=\fontsize{8}{9}\selectfont},
        livebar/.style={fill={rgb,255:red,156;green,164;blue,174}},
        pausedbar/.style={fill={rgb,255:red,65;green,111;blue,167}}]
        \foreach \tick in {0,20,40,60} {
            \draw[black!13,line width=0.35pt] (0,\tick/10) -- (4.25,\tick/10);
            \node[anchor=east,inner sep=3pt] at (0,\tick/10) {\tick};
        }
        \foreach \x/\height/\value in {0.49/0.600000/6.0,1.94/2.039312/20.4,3.39/4.516129/45.2} {
            \fill[livebar] (\x-0.19,0) rectangle (\x+0.19,\height);
            \node[anchor=south,inner sep=2pt] at (\x,\height) {\value};
        }
        \foreach \x/\height/\value in {0.91/2.980000/29.8,2.36/3.144963/31.4,3.81/5.268817/52.7} {
            \fill[pausedbar] (\x-0.19,0) rectangle (\x+0.19,\height);
            \node[anchor=south,inner sep=2pt] at (\x,\height) {\value};
        }
        \foreach \left/\right/\height in {0.49/0.91/4.30,1.94/2.36/4.45,3.39/3.81/6.60} {
            \draw[line width=0.45pt] (\left,\height-0.15) -- (\left,\height) -- (\right,\height) -- (\right,\height-0.15);
        }
        \node[anchor=south,font=\fontsize{7}{8}\selectfont,inner sep=2pt] at (0.70,4.30) {$p<0.001$};
        \node[anchor=south,font=\fontsize{7}{8}\selectfont,inner sep=2pt] at (2.15,4.45) {$p<0.001$};
        \node[anchor=south,font=\fontsize{7}{8}\selectfont,inner sep=2pt] at (3.60,6.60) {$p=0.443$};
        \draw[line width=0.5pt] (0,7.2) -- (0,0) -- (4.25,0);
        \node[rotate=90,font=\fontsize{9}{10}\selectfont] at (-0.48,3.3) {Pass rate (\%)};
        \node[anchor=north,align=center,inner sep=3pt,font=\fontsize{7}{8}\selectfont] at (0.70,0) {Real-time\\required};
        \node[anchor=north,align=center,inner sep=3pt,font=\fontsize{7}{8}\selectfont] at (2.15,0) {Non-real-time\\Temporal\\reasoning\\required};
        \node[anchor=north,align=center,inner sep=3pt,font=\fontsize{7}{8}\selectfont] at (3.60,0) {Non-real-time\\Temporal\\reasoning\\not required};
        \fill[livebar] (1.05,8.16) rectangle (1.30,8.44);
        \node[anchor=west,inner sep=3pt,font=\fontsize{9}{10}\selectfont] at (1.30,8.30) {Live};
        \fill[pausedbar] (2.20,8.16) rectangle (2.45,8.44);
        \node[anchor=west,inner sep=3pt,font=\fontsize{9}{10}\selectfont] at (2.45,8.30) {Paused};
    \end{tikzpicture}
    \end{minipage}\hfill
    \begin{minipage}[c]{0.48\textwidth}
    \caption{\textbf{Effect of pausing on Luna.} Pausing improves performance when real-time interaction or temporal reasoning is required. $p$-values are Holm-adjusted.}
    \label{fig:realtime-results}
    \end{minipage}
\end{figure}

Besides response speed, our controls let us study which other capabilities limit agents. We first examine whether the benefit of pausing is limited to real-time tasks. Figure~\ref{fig:realtime-results} compares live and paused execution for GPT-6 Luna with full interaction, averaged across all five difficulty levels. Interestingly, pausing also helps on the non-real-time counterparts that require temporal reasoning, raising Luna's pass rate from 20.4\% to 31.4\%, while the gain on counterparts without temporal reasoning, from 45.2\% to 52.7\%, is not statistically significant ($p=0.443$). Since these counterparts do not require acting in real time, this suggests that models also struggle to reason about scenes that change over time, even when they need not act quickly, identifying temporal reasoning as a further source of difficulty. Moreover, even when paused, Luna solves only 29.8\% of the real-time tasks, showing that for weaker models, response speed is only one of several bottlenecks.

\begin{figure}[t]
    \centering
    \tikzset{
        resultfont/.style={every node/.style={font=\fontsize{7}{8}\selectfont}},
        livestyle/.style={color={rgb,255:red,156;green,164;blue,174}},
        pausedstyle/.style={color={rgb,255:red,65;green,111;blue,167}},
        simplestyle/.style={color={rgb,255:red,74;green,145;blue,126}},
        fullstyle/.style={color={rgb,255:red,214;green,124;blue,58}}}
    \makebox[0.32\textwidth]{%
    \begin{tikzpicture}[baseline=0pt,x=0.82cm,y=0.026cm,resultfont]
        \foreach \t in {0,20,40,60,80,100} {
            \draw[black!13,line width=0.35pt] (0,\t) -- (4.4,\t);
            \node[anchor=east,inner sep=2pt] at (0,\t) {\t};
        }
        \draw[livestyle,line width=0.9pt] (0.4,92.86) -- (1.3,37.04) -- (2.2,22.12) -- (3.1,8.89) -- (4.0,7.20);
        \foreach \x/\y in {0.4/92.86,1.3/37.04,2.2/22.12,3.1/8.89,4.0/7.20} {
            \fill[livestyle] (\x-0.045,\y-1.7) rectangle (\x+0.045,\y+1.7);
        }
        \draw[pausedstyle,line width=0.9pt] (0.4,100) -- (1.3,55.56) -- (2.2,44.87) -- (3.1,22.83) -- (4.0,23.20);
        \foreach \x/\y in {0.4/100,1.3/55.56,2.2/44.87,3.1/22.83,4.0/23.20} {
            \fill[pausedstyle] (\x,\y) circle[radius=1.4pt];
        }
        \draw[line width=0.5pt] (0,108) -- (0,0) -- (4.4,0);
        \foreach \x/\l in {0.4/0,1.3/1,2.2/2,3.1/3,4.0/4} {
            \node[anchor=north,inner sep=2pt] at (\x,0) {\l};
        }
        \node at (2.2,-24) {Number of required capabilities};
        \node[rotate=90] at (-0.72,50) {Pass rate (\%)};
        \draw[livestyle,line width=0.9pt] (0.75,122) -- (1.05,122);
        \fill[livestyle] (0.855,120.3) rectangle (0.945,123.7);
        \node[anchor=west,inner sep=2pt] at (1.05,122) {Live};
        \draw[pausedstyle,line width=0.9pt] (2.25,122) -- (2.55,122);
        \fill[pausedstyle] (2.4,122) circle[radius=1.4pt];
        \node[anchor=west,inner sep=2pt] at (2.55,122) {Paused};
    \end{tikzpicture}}%
    \hfill
    \makebox[0.32\textwidth]{%
    \begin{tikzpicture}[baseline=0pt,x=1.2cm,y=0.13cm,resultfont]
        \foreach \t in {0,5,10,15,20} {
            \draw[black!13,line width=0.35pt] (0,\t) -- (3.0,\t);
            \node[anchor=east,inner sep=2pt] at (0,\t) {\t};
        }
        \fill[color={rgb,255:red,180;green,186;blue,210}] (0.42,0) rectangle (1.28,11.3);
        \fill[pausedstyle] (1.72,0) rectangle (2.58,16.2);
        \node[anchor=south,inner sep=2pt] at (0.85,11.3) {11.3};
        \node[anchor=south,inner sep=2pt] at (2.15,16.2) {16.2};
        \draw[line width=0.5pt] (0,21.6) -- (0,0) -- (3.0,0);
        \node[anchor=north,inner sep=2pt] at (0.85,0) {Without rules};
        \node[anchor=north,inner sep=2pt] at (2.15,0) {With rules};
        \node at (1.5,-4.8) {Interface rules};
        \node[rotate=90] at (-0.5,10) {Pass rate (\%)};
    \end{tikzpicture}}%
    \hfill
    \makebox[0.32\textwidth]{%
    \begin{tikzpicture}[baseline=0pt,x=0.72cm,y=0.0433cm,resultfont]
        \foreach \t in {0,20,40,60} {
            \draw[black!13,line width=0.35pt] (0,\t) -- (5.0,\t);
            \node[anchor=east,inner sep=2pt] at (0,\t) {\t};
        }
        \draw[simplestyle,line width=0.9pt] (0.5,53) -- (1.5,41) -- (2.5,30) -- (3.5,23) -- (4.5,19);
        \foreach \x/\y in {0.5/53,1.5/41,2.5/30,3.5/23,4.5/19} {
            \fill[simplestyle] (\x,\y) circle[radius=1.4pt];
        }
        \draw[fullstyle,line width=0.9pt] (0.5,53) -- (1.5,41) -- (2.5,29) -- (3.5,14) -- (4.5,12);
        \foreach \x/\y in {0.5/53,1.5/41,2.5/29,3.5/14,4.5/12} {
            \fill[fullstyle] (\x-0.05,\y-1.0) rectangle (\x+0.05,\y+1.0);
        }
        \foreach \x/\y in {0.5/53,1.5/41} {
            \node[anchor=south,inner sep=2.5pt] at (\x,\y) {\y};
        }
        \foreach \x/\y in {2.5/30,3.5/23,4.5/19} {
            \node[simplestyle,anchor=south,inner sep=2.5pt] at (\x,\y) {\y};
        }
        \foreach \x/\y in {2.5/29,3.5/14,4.5/12} {
            \node[fullstyle,anchor=north,inner sep=2.5pt] at (\x,\y) {\y};
        }
        \draw[line width=0.5pt] (0,65) -- (0,0) -- (5.0,0);
        \foreach \x/\l in {0.5/1,1.5/2,2.5/3,3.5/4,4.5/5} {
            \node[anchor=north,inner sep=2pt] at (\x,0) {\l};
        }
        \node at (2.5,-14.4) {Difficulty};
        \node[rotate=90] at (-0.8,30) {Pass rate (\%)};
        \draw[simplestyle,line width=0.9pt] (0.35,73) -- (0.7,73);
        \fill[simplestyle] (0.525,73) circle[radius=1.4pt];
        \node[anchor=west,inner sep=2pt] at (0.7,73) {Simplified};
        \draw[fullstyle,line width=0.9pt] (2.75,73) -- (3.1,73);
        \fill[fullstyle] (2.875,72) rectangle (2.975,74);
        \node[anchor=west,inner sep=2pt] at (3.1,73) {Full};
    \end{tikzpicture}}

    \vspace{2pt}
    \begin{minipage}[t]{0.32\textwidth}
        \caption{\textbf{Performance with increasing capability requirements.} Luna's pass rate generally decreases as more of the four task-level capabilities are required.}
        \label{fig:capability-results}
    \end{minipage}\hfill
    \begin{minipage}[t]{0.32\textwidth}
        \caption{\textbf{Effect of providing interface rules.} Luna solves more tasks when given the rules, across real-time and non-real-time environments.}
        \label{fig:exploration-results}
    \end{minipage}\hfill
    \begin{minipage}[t]{0.32\textwidth}
        \caption{\textbf{Effect of difficulty and interaction.} Under paused execution, simplified controls improve Luna's performance at higher difficulties on real-time tasks.}
        \label{fig:difficulty-results}
    \end{minipage}
\end{figure}

\subsection{Performance Decreases as More Capabilities Are Required}

Since review agents label the capabilities each configuration requires, we can also relate performance to the demands of a task. Figure~\ref{fig:capability-results} reports Luna's pass rate as a function of the number of required capabilities, using full interaction across all five difficulty levels. In the live setting, pass rate decreases from 37.0\% when one capability is required to 7.2\% when all four are required, and paused execution shows the same trend, from 55.6\% to 23.2\%. Since the decline persists when the environment is paused, tasks become harder as capabilities are combined, even without time pressure. However, since capability requirements overlap across tasks, this analysis cannot isolate the effect of a single capability. We therefore examine interface exploration directly through a controlled experiment in Section~\ref{sec:exploration-results}.

\subsection{Providing Interface Rules Improves Performance}
\label{sec:exploration-results}

Many minigames require the agent to learn how an unfamiliar interface behaves before it can solve the task. To isolate this demand, we append an environment-specific rulebook describing the controls, action effects, and world rules to Luna's task prompt (Appendix~\ref{app:rulebooks}), and evaluate with full interaction in the live setting on minigames requiring exploration and interface understanding. Figure~\ref{fig:exploration-results} shows that providing these rules raises Luna's pass rate from 11.3\% to 16.2\% across the five difficulty levels. The model must still solve the same minigame and carry out its solution through the same controls, so this gain comes from knowing the rules in advance. This identifies interface exploration as a further source of difficulty: Luna can use interface knowledge that it struggles to acquire through interaction.

\subsection{Performance Decreases with Difficulty and Improves with Simplified Interaction}
\label{sec:difficulty-results}

\begin{table}[t]
    \centering
    \caption{\textbf{Pass rates (\%) of GPT-6 Luna (xHigh) across difficulty levels, with full interaction in the live setting.} Gap is the pass rate on the non-real-time counterparts minus the pass rate on the real-time tasks.}
    \label{tab:difficulty-live}
    \small
    \begin{tabular}{@{}lrrrrrr@{}}
        \toprule
        & \multicolumn{5}{c}{\textbf{Difficulty}} & \\
        \cmidrule(lr){2-6}
        & \textbf{1} & \textbf{2} & \textbf{3} & \textbf{4} & \textbf{5} & \textbf{All} \\
        \midrule
        Real-time & 15.0 & 6.0 & 4.0 & 4.0 & 1.0 & 6.0 \\
        Non-real-time & 35.0 & 33.0 & 22.0 & 18.0 & 17.0 & 25.0 \\
        Gap & 20.0 & 27.0 & 18.0 & 14.0 & 16.0 & 19.0 \\
        \bottomrule
    \end{tabular}
\end{table}

Finally, we turn to the easier difficulty levels and simplified interaction. Table~\ref{tab:difficulty-live} reports Luna's pass rates in the live setting with full interaction across all five difficulty levels. While Luna solves only 1.0\% of the real-time tasks at the highest difficulty, it solves 15.0\% at difficulty 1, so the easier levels leave room to measure progress for weaker models. On their non-real-time counterparts, its pass rate falls from 35.0\% to 17.0\%. The gap between the two stays between 14.0 and 27.0 points at every difficulty, so acting in real time is a major source of failure even on the easiest tasks. To study difficulty without this effect, Figure~\ref{fig:difficulty-results} reports Luna's pass rate on real-time tasks across the five difficulty levels under paused execution. Averaged across both interaction modes, pass rate decreases from 53.0\% at difficulty 1 to 15.5\% at difficulty 5, showing that the difficulty levels provide a controlled range of challenge. Simplified interaction raises the overall pass rate from 29.8\% to 33.2\%, with larger gains at higher difficulty. Since both modes realize the same minigame, this gain shows that some failures arise from operating the full controls rather than from solving the task. However, the decline with difficulty persists under simplified interaction, showing that the challenges introduced at higher difficulty levels are not only about the controls.

\FloatBarrier

\section{Related Work}
\label{sec:related-work}

\textbf{Real-time interaction.}
Acting in a changing environment requires accounting for the delay between observing it and acting on it. Robotics methods handle this delay through asynchronous execution or by predicting the state at execution time \citep{black2025realtimeexecutionactionchunking,shukor2025smolvla,tang2025vlash}, and reinforcement learning studies decision making under action and observation delays \citep{ramstedt2019realtime,derman2021acting}. For computer-use agents, OSWorld-Human measures latency and finds that agents take 2.7 to 4.3 times more steps than necessary \citep{abhyankar2026osworldhumanbenchmarkingefficiencycomputeruse}. Game benchmarks either pause the game during inference \citep{ouyang2026gameworldstandardizedverifiableevaluation} or compare paused and real-time play \citep{zhang2026videogamebenchvisionlanguagemodelscomplete,park2025orak}, and Real-Time Reasoning Gym adds time pressure to text-based games \citep{wen2026realtime}. However, these comparisons cover only a few visual games or text-based environments. \bench{} pairs each of 100 real-time minigames with a non-real-time counterpart and controls the clock of each minigame, so that we can vary how much time an agent has to respond on the same task.

\textbf{Benchmarks for interactive visual agents.}
Benchmarks for vision-language agents cover websites \citep{zhou2024webarena,koh2024visualwebarena}, desktop and mobile applications \citep{xie2024osworldbenchmarkingmultimodalagents,bonatti2024windowsagentarena,yuan2026osworld2,rawles2024androidworld}, professional software \citep{aggarwal2025programmingpixelscomputeruseagents,aggarwal2026gymanythingturnsoftwareagent}, games \citep{zhang2026videogamebenchvisionlanguagemodelscomplete,paglieri2025balrog,hu2025lmgamebench,zheng2025vmage,ying2026aigamestorescalableopenended,ouyang2026gameworldstandardizedverifiableevaluation,arcprize2026arcagi3}, and embodied simulators \citep{savva2019habitatplatformembodiedai,Shridhar_2020_CVPR,srivastava2021behaviorbenchmarkeverydayhousehold,yang2025embodiedbenchcomprehensivebenchmarkingmultimodal}. However, most of these environments wait for the agent between actions, and each task requires many capabilities at once, so low performance does not reveal which capability limits the agent. Prior works address the latter by mapping tasks to capabilities \citep{wu2023smartplay}, evaluating perception, reasoning, and decision making separately \citep{wang2025lvlmplayground,xu2026vsbench,li2024embodiedagentinterface}, or varying one condition of a task, such as difficulty, the initial state, or the interface \citep{davydova2025osuniverse,zheng2025mcu,zhao2025worldgui,liu2026dynamicguibench}. \bench{} instead varies the real-time requirement, difficulty, and interaction interface of the same minigame.

\textbf{Capability-specific evaluation of VLMs.}
Many benchmarks test individual capabilities of VLMs through questions about images or videos, such as 3D spatial understanding \citep{Yang_2025_CVPR,wang2025mindcube,ma2024threedsrbench,jia2025omnispatial}, temporal reasoning \citep{liu-etal-2024-tempcompass,shangguan2024tomato,zhang2024vinoground}, GUI grounding \citep{cheng-etal-2024-seeclick,li2025screenspot}, and abstract reasoning \citep{chia2024puzzlevqa,ghosal2024language,song2025visualpuzzles}. However, the model answers a question about a fixed input, without acting on what it perceives. iVISPAR and VisGym evaluate visual reasoning through interaction \citep{mayer2025ivispar,wang2026visgym}. \bench{} further requires agents to perceive and act while the minigame keeps changing.

\textbf{Automated environment construction.}
Recent works use coding agents and LLMs to construct computer-use environments \citep{aggarwal2026gymanythingturnsoftwareagent,wang2026cuagym,cao2026guigenesis}, text-based environments \citep{zhang2025autoenv}, and environments with tasks of varying difficulty \citep{hu2024agentgen,faldor2024omniepic}, while procedural generators vary the complexity of task instances \citep{cobbe2020leveraging,samvelyan2021minihack,chevalierboisvert2019babyai,stojanovski2025reasoning,chen2025enigmata}. However, none of these produce visual minigames whose timing, difficulty, and interaction interface can be varied while preserving the underlying task. We extend the creator and audit agents of Gym-Anything \citep{aggarwal2026gymanythingturnsoftwareagent} to construct such minigames, each with a non-real-time counterpart and an oracle solution. Appendix~\ref{app:related-work} provides a more detailed related work.

\section{Conclusion}
\label{sec:conclusion}

In this work, we introduced \bench{}, a scalable framework for creating environments for studying the real-time interaction capabilities of vision-language agents. In \bench{}, coding agents implement and audit each environment as a visual minigame program with an oracle solution, and our evaluation stack controls the clock of each minigame. As a result, we can vary how much time an agent has to respond, whether the task requires real-time interaction, its difficulty, and its interaction interface on the same minigame. Using \bench{}, we constructed a benchmark of 100 real-time minigames and showed that current agents perform poorly in real time, with the strongest agent, GPT-6 Astra (xHigh), solving only 9.0\% of the real-time tasks compared with 66.0\% of their non-real-time counterparts. We further showed that speed is the primary bottleneck for frontier models, and that temporal reasoning and exploration of unfamiliar interfaces are additional sources of difficulty. Interestingly, we also find that frontier agents can partially offset their poor real-time capabilities by writing code that perceives and acts on their behalf.

More broadly, our results suggest that progress in real-time interaction requires agents that are not only capable but also fast. We hope that \bench{} and the released minigames provide a foundation for future work on real-time agents, including faster models, agents that move the time-critical parts of a task into code, and training on real-time environments.

\section*{Acknowledgements}

We thank Daytona for their generous support in sandbox compute credits. We thank Jing Yu Koh, Lawrence Jang, Seungone Kim, Zeyu Zheng, and other members of the L3 Lab for feedback and helpful discussions. Pranjal Aggarwal is supported by a SoftBank Group-Arm Fellowship. This work is partially supported by the National Science Foundation under Grant No. DMS-2502281.

\FloatBarrier
\bibliography{references}

\begin{thebibliography}{77}
\providecommand{\natexlab}[1]{#1}
\providecommand{\url}[1]{\texttt{#1}}
\expandafter\ifx\csname urlstyle\endcsname\relax
  \providecommand{\doi}[1]{doi: #1}\else
  \providecommand{\doi}{doi: \begingroup \urlstyle{rm}\Url}\fi

\bibitem[Abhyankar et~al.(2026)Abhyankar, Qi, and
  Zhang]{abhyankar2026osworldhumanbenchmarkingefficiencycomputeruse}
Reyna Abhyankar, Qi~Qi, and Yiying Zhang.
\newblock {OSWorld-Human: Benchmarking the Efficiency of Computer-Use Agents},
  2026.
\newblock URL \url{https://arxiv.org/abs/2506.16042}.

\bibitem[Aggarwal \& Welleck(2025)Aggarwal and
  Welleck]{aggarwal2025programmingpixelscomputeruseagents}
Pranjal Aggarwal and Sean Welleck.
\newblock {Programming with Pixels: Can Computer-Use Agents do Software
  Engineering?}, 2025.
\newblock URL \url{https://arxiv.org/abs/2502.18525}.

\bibitem[Aggarwal et~al.(2026)Aggarwal, Neubig, and
  Welleck]{aggarwal2026gymanythingturnsoftwareagent}
Pranjal Aggarwal, Graham Neubig, and Sean Welleck.
\newblock {Gym-Anything: Turn any Software into an Agent Environment}, 2026.
\newblock URL \url{https://arxiv.org/abs/2604.06126}.

\bibitem[{ARC Prize Foundation}(2026)]{arcprize2026arcagi3}
{ARC Prize Foundation}.
\newblock {{ARC-AGI-3}: A New Challenge for Frontier Agentic Intelligence},
  2026.

\bibitem[Battleday et~al.(2026)Battleday, Sandbrink, Cullen-Drohan, Yan,
  Muller, Maguire, Kubicek, Greenlee-Scott, Sumant, Dao, Schmidhuber, Valko,
  Tenenbaum, Griffiths, Kurth-Nelson, and Whittington]{battleday2026dig}
Ruairidh~M. Battleday, Kai Sandbrink, Jimi Cullen-Drohan, Zihan Yan, Timothy
  Muller, Clare Maguire, Ales Kubicek, Fraser Greenlee-Scott, Sukrit Sumant,
  Tri Dao, J{\"u}rgen Schmidhuber, Michal Valko, Joshua Tenenbaum, Thomas~L.
  Griffiths, Zeb Kurth-Nelson, and James C.~R. Whittington.
\newblock {{DiG}-bench: Discovery in Games}, 2026.
\newblock URL \url{https://arxiv.org/abs/2608.12593}.

\bibitem[Black et~al.(2025)Black, Galliker, and
  Levine]{black2025realtimeexecutionactionchunking}
Kevin Black, Manuel~Y. Galliker, and Sergey Levine.
\newblock {Real-Time Execution of Action Chunking Flow Policies}, 2025.
\newblock URL \url{https://arxiv.org/abs/2506.07339}.

\bibitem[Bonatti et~al.(2024)Bonatti, Zhao, Bonacci, Dupont, Abdali, Li, Lu,
  Wagle, Koishida, Bucker, Jang, and Hui]{bonatti2024windowsagentarena}
Rogerio Bonatti, Dan Zhao, Francesco Bonacci, Dillon Dupont, Sara Abdali,
  Yinheng Li, Yadong Lu, Justin Wagle, Kazuhito Koishida, Arthur Bucker,
  Lawrence Jang, and Zack Hui.
\newblock {Windows Agent Arena: Evaluating Multi-Modal {OS} Agents at Scale},
  2024.

\bibitem[Cao et~al.(2026)Cao, Ran, Wu, Guo, Chen, Li, Cao, Zhi, Yu, Li, Yang,
  and Xie]{cao2026guigenesis}
Yuan Cao, Dezhi Ran, Mengzhou Wu, Yuzhe Guo, Xin Chen, Ang Li, Gang Cao, Gong
  Zhi, Hao Yu, Linyi Li, Wei Yang, and Tao Xie.
\newblock {GUI-GENESIS: Automated Synthesis of Efficient Environments with
  Verifiable Rewards for GUI Agent Post-Training}, 2026.
\newblock URL \url{https://arxiv.org/abs/2602.14093}.

\bibitem[Chen et~al.(2025)Chen, He, Yuan, Chen, Cai, Dai, Yu, Yu, Li, Chen,
  Zhou, and Wang]{chen2025enigmata}
Jiangjie Chen, Qianyu He, Siyu Yuan, Aili Chen, Zhicheng Cai, Weinan Dai,
  Hongli Yu, Qiying Yu, Xuefeng Li, Jiaze Chen, Hao Zhou, and Mingxuan Wang.
\newblock {Enigmata: Scaling Logical Reasoning in Large Language Models with
  Synthetic Verifiable Puzzles}, 2025.
\newblock URL \url{https://arxiv.org/abs/2505.19914}.

\bibitem[Chen et~al.(2026)Chen, Zhai, Wang, Yang, Zhang, Wang, and
  Zhang]{chen2026captcha}
Yuxi Chen, Haoyu Zhai, Chenkai Wang, Rui Yang, Lingming Zhang, Gang Wang, and
  Huan Zhang.
\newblock {CAPTCHA Solving for Native GUI Agents: Automated Reasoning-Action
  Data Generation and Self-Corrective Training}.
\newblock In \emph{International Conference on Machine Learning (ICML)}, 2026.
\newblock arXiv:2603.23559.

\bibitem[Cheng et~al.(2024)Cheng, Sun, Chu, Xu, YanTao, Zhang, and
  Wu]{cheng-etal-2024-seeclick}
Kanzhi Cheng, Qiushi Sun, Yougang Chu, Fangzhi Xu, Li~YanTao, Jianbing Zhang,
  and Zhiyong Wu.
\newblock {{S}ee{C}lick: Harnessing {GUI} Grounding for Advanced Visual {GUI}
  Agents}.
\newblock In Lun-Wei Ku, Andre Martins, and Vivek Srikumar (eds.),
  \emph{Proceedings of the 62nd Annual Meeting of the Association for
  Computational Linguistics (Volume 1: Long Papers)}, pp.\  9313--9332,
  Bangkok, Thailand, August 2024. Association for Computational Linguistics.
\newblock \doi{10.18653/v1/2024.acl-long.505}.
\newblock URL \url{https://aclanthology.org/2024.acl-long.505/}.

\bibitem[Chevalier-Boisvert et~al.(2019)Chevalier-Boisvert, Bahdanau, Lahlou,
  Willems, Saharia, Nguyen, and Bengio]{chevalierboisvert2019babyai}
Maxime Chevalier-Boisvert, Dzmitry Bahdanau, Salem Lahlou, Lucas Willems,
  Chitwan Saharia, Thien~Huu Nguyen, and Yoshua Bengio.
\newblock {BabyAI: A Platform to Study the Sample Efficiency of Grounded
  Language Learning}.
\newblock In \emph{International Conference on Learning Representations
  (ICLR)}, 2019.

\bibitem[Chia et~al.(2024)Chia, Han, Ghosal, Bing, and
  Poria]{chia2024puzzlevqa}
Yew~Ken Chia, Vernon Toh~Yan Han, Deepanway Ghosal, Lidong Bing, and Soujanya
  Poria.
\newblock {PuzzleVQA: Diagnosing Multimodal Reasoning Challenges of Language
  Models with Abstract Visual Patterns}, 2024.
\newblock URL \url{https://arxiv.org/abs/2403.13315}.

\bibitem[Cobbe et~al.(2019)Cobbe, Hesse, Hilton, and
  Schulman]{cobbe2020leveraging}
Karl Cobbe, Christopher Hesse, Jacob Hilton, and John Schulman.
\newblock {Leveraging Procedural Generation to Benchmark Reinforcement
  Learning}, 2019.
\newblock URL \url{https://arxiv.org/abs/1912.01588}.

\bibitem[Davydova et~al.(2025)Davydova, Jeffries, Barker, M{\'a}rquez~Flores,
  and Ryan]{davydova2025osuniverse}
Mariya Davydova, Daniel Jeffries, Patrick Barker, Arturo M{\'a}rquez~Flores,
  and Sin{\'e}ad Ryan.
\newblock {{OSUniverse}: Benchmark for Multimodal {GUI}-navigation {AI}
  Agents}, 2025.

\bibitem[{DeepSeek-AI}(2025)]{deepseekai2025deepseekr1}
{DeepSeek-AI}.
\newblock {{DeepSeek-R1}: Incentivizing Reasoning Capability in {LLMs} via
  Reinforcement Learning}.
\newblock \emph{arXiv preprint arXiv:2501.12948}, 2025.

\bibitem[Deng et~al.(2024)Deng, Ou, Liu, Zhang, Zhang, and
  Liu]{deng2024oedipus}
Gelei Deng, Haoran Ou, Yi~Liu, Jie Zhang, Tianwei Zhang, and Yang Liu.
\newblock {Oedipus: LLM-enchanced Reasoning CAPTCHA Solver}, 2024.
\newblock URL \url{https://arxiv.org/abs/2405.07496}.

\bibitem[Derman et~al.(2021)Derman, Dalal, and Mannor]{derman2021acting}
Esther Derman, Gal Dalal, and Shie Mannor.
\newblock {Acting in Delayed Environments with Non-Stationary Markov Policies}.
\newblock In \emph{International Conference on Learning Representations
  (ICLR)}, 2021.
\newblock arXiv:2101.11992.

\bibitem[Faldor et~al.(2024)Faldor, Zhang, Cully, and
  Clune]{faldor2024omniepic}
Maxence Faldor, Jenny Zhang, Antoine Cully, and Jeff Clune.
\newblock {OMNI-EPIC: Open-endedness via Models of human Notions of
  Interestingness with Environments Programmed in Code}, 2024.
\newblock URL \url{https://arxiv.org/abs/2405.15568}.

\bibitem[Fu et~al.(2024)Fu, Hu, Li, Feng, Wang, Lin, Roth, Smith, Ma, and
  Krishna]{fu2024blink}
Xingyu Fu, Yushi Hu, Bangzheng Li, Yu~Feng, Haoyu Wang, Xudong Lin, Dan Roth,
  Noah~A. Smith, Wei-Chiu Ma, and Ranjay Krishna.
\newblock {BLINK: Multimodal Large Language Models Can See but Not Perceive},
  2024.
\newblock URL \url{https://arxiv.org/abs/2404.12390}.

\bibitem[Ghosal et~al.(2024)Ghosal, Han, Ken, and Poria]{ghosal2024language}
Deepanway Ghosal, Vernon Toh~Yan Han, Chia~Yew Ken, and Soujanya Poria.
\newblock {Are Language Models Puzzle Prodigies? Algorithmic Puzzles Unveil
  Serious Challenges in Multimodal Reasoning}, 2024.
\newblock URL \url{https://arxiv.org/abs/2403.03864}.

\bibitem[Hu et~al.(2025{\natexlab{a}})Hu, Huo, Zhang, Yu, Xing, Stoica, Rosing,
  Jin, and Zhang]{hu2025lmgamebench}
Lanxiang Hu, Mingjia Huo, Yuxuan Zhang, Haoyang Yu, Eric~P. Xing, Ion Stoica,
  Tajana Rosing, Haojian Jin, and Hao Zhang.
\newblock {lmgame-Bench: How Good are {LLMs} at Playing Games?},
  2025{\natexlab{a}}.

\bibitem[Hu et~al.(2025{\natexlab{b}})Hu, Zhao, Xu, Sun, Lou, Lin, Luo, and
  Rajmohan]{hu2024agentgen}
Mengkang Hu, Pu~Zhao, Can Xu, Qingfeng Sun, Jianguang Lou, Qingwei Lin, Ping
  Luo, and Saravan Rajmohan.
\newblock {AgentGen: Enhancing Planning Abilities for Large Language Model
  based Agent via Environment and Task Generation}.
\newblock In \emph{Proceedings of the ACM SIGKDD Conference on Knowledge
  Discovery and Data Mining (KDD)}, 2025{\natexlab{b}}.

\bibitem[Jang et~al.(2026)Jang, Koh, Fried, and
  Salakhutdinov]{jang2026odysseys}
Lawrence~Keunho Jang, Jing~Yu Koh, Daniel Fried, and Ruslan Salakhutdinov.
\newblock {Odysseys: Benchmarking Web Agents on Realistic Long Horizon Tasks},
  2026.
\newblock URL \url{https://arxiv.org/abs/2604.24964}.

\bibitem[Jia et~al.(2025)Jia, Qi, Zhang, Zhang, Yu, He, Wang, and
  Yi]{jia2025omnispatial}
Mengdi Jia, Zekun Qi, Shaochen Zhang, Wenyao Zhang, Xinqiang Yu, Jiawei He,
  He~Wang, and Li~Yi.
\newblock {OmniSpatial: Towards Comprehensive Spatial Reasoning Benchmark for
  Vision Language Models}, 2025.
\newblock URL \url{https://arxiv.org/abs/2506.03135}.

\bibitem[Jimenez et~al.(2024)Jimenez, Yang, Wettig, Yao, Pei, Press, and
  Narasimhan]{jimenez2024swebench}
Carlos~E. Jimenez, John Yang, Alexander Wettig, Shunyu Yao, Kexin Pei, Ofir
  Press, and Karthik Narasimhan.
\newblock {{SWE}-bench: Can Language Models Resolve Real-World {GitHub}
  Issues?}
\newblock In \emph{The Twelfth International Conference on Learning
  Representations}, 2024.

\bibitem[Koh et~al.(2024)Koh, Lo, Jang, Duvvur, Lim, Huang, Neubig, Zhou,
  Salakhutdinov, and Fried]{koh2024visualwebarena}
Jing~Yu Koh, Robert Lo, Lawrence Jang, Vikram Duvvur, Ming~Chong Lim, Po-Yu
  Huang, Graham Neubig, Shuyan Zhou, Ruslan Salakhutdinov, and Daniel Fried.
\newblock {{VisualWebArena}: Evaluating Multimodal Agents on Realistic Visual
  Web Tasks}.
\newblock In \emph{Proceedings of the Annual Meeting of the Association for
  Computational Linguistics (ACL)}, 2024.
\newblock arXiv:2401.13649.

\bibitem[Li et~al.(2025)Li, Meng, Lin, Luo, Tian, Ma, Huang, and
  Chua]{li2025screenspot}
Kaixin Li, Ziyang Meng, Hongzhan Lin, Ziyang Luo, Yuchen Tian, Jing Ma, Zhiyong
  Huang, and Tat-Seng Chua.
\newblock {ScreenSpot-Pro: GUI Grounding for Professional High-Resolution
  Computer Use}, 2025.
\newblock URL \url{https://arxiv.org/abs/2504.07981}.

\bibitem[Li et~al.(2024)Li, Zhao, Wang, Wang, Zhou, Srivastava, Gokmen, Lee,
  Li, Zhang, Liu, Liang, Fei-Fei, Mao, and Wu]{li2024embodiedagentinterface}
Manling Li, Shiyu Zhao, Qineng Wang, Kangrui Wang, Yu~Zhou, Sanjana Srivastava,
  Cem Gokmen, Tony Lee, Li~Erran Li, Ruohan Zhang, Weiyu Liu, Percy Liang,
  Li~Fei-Fei, Jiayuan Mao, and Jiajun Wu.
\newblock {Embodied Agent Interface: Benchmarking {LLMs} for Embodied Decision
  Making}.
\newblock In \emph{Advances in Neural Information Processing Systems (NeurIPS)
  Datasets and Benchmarks Track}, 2024.
\newblock arXiv:2410.07166.

\bibitem[Liu et~al.(2026)Liu, Pan, Gao, Yang, Shi, Liu, Wu, and
  Li]{liu2026dynamicguibench}
Enqi Liu, Liyuan Pan, Zhi Gao, Yan Yang, Chenrui Shi, Yang Liu, Jingrong Wu,
  and Qing Li.
\newblock {Benchmarking and Improving {GUI} Agents in High-Dynamic
  Environments}, 2026.

\bibitem[Liu et~al.(2024)Liu, Li, Liu, Wang, Ren, Li, Chen, Sun, and
  Hou]{liu-etal-2024-tempcompass}
Yuanxin Liu, Shicheng Li, Yi~Liu, Yuxiang Wang, Shuhuai Ren, Lei Li, Sishuo
  Chen, Xu~Sun, and Lu~Hou.
\newblock {{T}emp{C}ompass: Do Video {LLM}s Really Understand Videos?}
\newblock In Lun-Wei Ku, Andre Martins, and Vivek Srikumar (eds.),
  \emph{Findings of the Association for Computational Linguistics: ACL 2024},
  pp.\  8731--8772, Bangkok, Thailand, August 2024. Association for
  Computational Linguistics.
\newblock \doi{10.18653/v1/2024.findings-acl.517}.
\newblock URL \url{https://aclanthology.org/2024.findings-acl.517/}.

\bibitem[Luo et~al.(2025)Luo, Li, Liu, Cui, Zhao, and
  Shen]{luo2025opencaptchaworld}
Yaxin Luo, Zhaoyi Li, Jiacheng Liu, Jiacheng Cui, Xiaohan Zhao, and Zhiqiang
  Shen.
\newblock {Open CaptchaWorld: A Comprehensive Web-based Platform for Testing
  and Benchmarking Multimodal LLM Agents}, 2025.
\newblock URL \url{https://arxiv.org/abs/2505.24878}.

\bibitem[Ma et~al.(2024)Ma, Chen, Zhang, Chou, Chen, de~Melo, and
  Yuille]{ma2024threedsrbench}
Wufei Ma, Haoyu Chen, Guofeng Zhang, Yu-Cheng Chou, Jieneng Chen, Celso~M
  de~Melo, and Alan Yuille.
\newblock {3DSRBench: A Comprehensive 3D Spatial Reasoning Benchmark}, 2024.
\newblock URL \url{https://arxiv.org/abs/2412.07825}.

\bibitem[Mayer et~al.(2025)Mayer, Ballout, Jassim, Nosrat~Nezami, and
  Bruni]{mayer2025ivispar}
Julius Mayer, Mohamad Ballout, Serwan Jassim, Farbod Nosrat~Nezami, and Elia
  Bruni.
\newblock {{iVISPAR} -- An Interactive Visual-Spatial Reasoning Benchmark for
  {VLMs}}, 2025.

\bibitem[Ouyang et~al.(2026)Ouyang, Hu, Lin, Ng, and
  Shou]{ouyang2026gameworldstandardizedverifiableevaluation}
Mingyu Ouyang, Siyuan Hu, Kevin~Qinghong Lin, Hwee~Tou Ng, and Mike~Zheng Shou.
\newblock {GameWorld: Towards Standardized and Verifiable Evaluation of
  Multimodal Game Agents}, 2026.
\newblock URL \url{https://arxiv.org/abs/2604.07429}.

\bibitem[Paglieri et~al.(2025)Paglieri, Cupia{\l}, Coward, Piterbarg, Wolczyk,
  Khan, Pignatelli, Kuci{\'n}ski, Pinto, Fergus, Foerster, Parker-Holder, and
  Rockt{\"a}schel]{paglieri2025balrog}
Davide Paglieri, Bart{\l}omiej Cupia{\l}, Samuel Coward, Ulyana Piterbarg,
  Maciej Wolczyk, Akbir Khan, Eduardo Pignatelli, {\L}ukasz Kuci{\'n}ski,
  Lerrel Pinto, Rob Fergus, Jakob~Nicolaus Foerster, Jack Parker-Holder, and
  Tim Rockt{\"a}schel.
\newblock {{BALROG}: Benchmarking Agentic {LLM} and {VLM} Reasoning On Games}.
\newblock In \emph{International Conference on Learning Representations
  (ICLR)}, 2025.
\newblock arXiv:2411.13543.

\bibitem[Pan et~al.(2025)Pan, Wang, Neubig, Jaitly, Ji, Suhr, and
  Zhang]{pan2025swegym}
Jiayi Pan, Xingyao Wang, Graham Neubig, Navdeep Jaitly, Heng Ji, Alane Suhr,
  and Yizhe Zhang.
\newblock {Training Software Engineering Agents and Verifiers with SWE-Gym}.
\newblock In \emph{Proceedings of the International Conference on Machine
  Learning (ICML)}, 2025.

\bibitem[Park et~al.(2025)Park, Kim, Choi, Kim, Lee, Lee, Park, Lee, Hwang,
  Ahn, et~al.]{park2025orak}
Dongmin Park, Minkyu Kim, Beongjun Choi, Junhyuck Kim, Keon Lee, Jonghyun Lee,
  Inkyu Park, Byeong-Uk Lee, Jaeyoung Hwang, Jaewoo Ahn, et~al.
\newblock {Orak: A Foundational Benchmark for Training and Evaluating {LLM}
  Agents on Diverse Video Games}, 2025.

\bibitem[Ramstedt \& Pal(2019)Ramstedt and Pal]{ramstedt2019realtime}
Simon Ramstedt and Christopher Pal.
\newblock {Real-Time Reinforcement Learning}.
\newblock In \emph{Advances in Neural Information Processing Systems
  (NeurIPS)}, 2019.
\newblock arXiv:1911.04448.

\bibitem[Rawles et~al.(2024)Rawles, Clinckemaillie, Chang, Waltz, Lau, Fair,
  Li, Bishop, Li, Campbell-Ajala, Toyama, Berry, Tyamagundlu, Lillicrap, and
  Riva]{rawles2024androidworld}
Christopher Rawles, Sarah Clinckemaillie, Yifan Chang, Jonathan Waltz,
  Gabrielle Lau, Marybeth Fair, Alice Li, William Bishop, Wei Li, Folawiyo
  Campbell-Ajala, Daniel Toyama, Robert Berry, Divya Tyamagundlu, Timothy
  Lillicrap, and Oriana Riva.
\newblock {{AndroidWorld}: A Dynamic Benchmarking Environment for Autonomous
  Agents}, 2024.

\bibitem[Ren et~al.(2025)Ren, Tertikas, Maiti, Han, Zhang, S{\"u}sstrunk, and
  Kokkinos]{ren2025vgrpbench}
Yufan Ren, Konstantinos Tertikas, Shalini Maiti, Junlin Han, Tong Zhang, Sabine
  S{\"u}sstrunk, and Filippos Kokkinos.
\newblock {{VGRP-Bench}: Visual Grid Reasoning Puzzle Benchmark for Large
  Vision-Language Models}, 2025.
\newblock URL \url{https://arxiv.org/abs/2503.23064}.

\bibitem[Samvelyan et~al.(2021)Samvelyan, Kirk, Kurin, Parker-Holder, Jiang,
  Hambro, Petroni, Küttler, Grefenstette, and
  Rocktäschel]{samvelyan2021minihack}
Mikayel Samvelyan, Robert Kirk, Vitaly Kurin, Jack Parker-Holder, Minqi Jiang,
  Eric Hambro, Fabio Petroni, Heinrich Küttler, Edward Grefenstette, and Tim
  Rocktäschel.
\newblock {MiniHack the Planet: A Sandbox for Open-Ended Reinforcement Learning
  Research}.
\newblock In \emph{Advances in Neural Information Processing Systems Datasets
  and Benchmarks Track}, 2021.

\bibitem[Savva et~al.(2019)Savva, Kadian, Maksymets, Zhao, Wijmans, Jain,
  Straub, Liu, Koltun, Malik, Parikh, and
  Batra]{savva2019habitatplatformembodiedai}
Manolis Savva, Abhishek Kadian, Oleksandr Maksymets, Yili Zhao, Erik Wijmans,
  Bhavana Jain, Julian Straub, Jia Liu, Vladlen Koltun, Jitendra Malik, Devi
  Parikh, and Dhruv Batra.
\newblock {Habitat: A Platform for Embodied AI Research}, 2019.
\newblock URL \url{https://arxiv.org/abs/1904.01201}.

\bibitem[Shangguan et~al.(2024)Shangguan, Li, Ding, Zheng, Zhao, Fitzgerald,
  and Cohan]{shangguan2024tomato}
Ziyao Shangguan, Chuhan Li, Yuxuan Ding, Yanan Zheng, Yilun Zhao, Tesca
  Fitzgerald, and Arman Cohan.
\newblock {TOMATO: Assessing Visual Temporal Reasoning Capabilities in
  Multimodal Foundation Models}, 2024.
\newblock URL \url{https://arxiv.org/abs/2410.23266}.

\bibitem[Shi et~al.(2025)Shi, Yang, Liu, Bu, Chen, Zhou, Ma, Wen, Wang, He,
  Song, Zhu, Li, Wang, Zhang, Yuan, Yao, Yang, Wang, Fang, Yuan, He, Tang, Tan,
  Zhou, Zhang, Li, Huang, and Zhang]{shi2025korgym}
Jiajun Shi, Jian Yang, Jiaheng Liu, Xingyuan Bu, Jiangjie Chen, Junting Zhou,
  Kaijing Ma, Zhoufutu Wen, Bingli Wang, Yancheng He, Liang Song, Hualei Zhu,
  Shilong Li, Xingjian Wang, Wei Zhang, Ruibin Yuan, Yifan Yao, Wenjun Yang,
  Yunli Wang, Siyuan Fang, Siyu Yuan, Qianyu He, Xiangru Tang, Yingshui Tan,
  Wangchunshu Zhou, Zhaoxiang Zhang, Zhoujun Li, Wenhao Huang, and Ge~Zhang.
\newblock {KORGym: A Dynamic Game Platform for LLM Reasoning Evaluation}, 2025.
\newblock URL \url{https://arxiv.org/abs/2505.14552}.

\bibitem[Shridhar et~al.(2020)Shridhar, Thomason, Gordon, Bisk, Han, Mottaghi,
  Zettlemoyer, and Fox]{Shridhar_2020_CVPR}
Mohit Shridhar, Jesse Thomason, Daniel Gordon, Yonatan Bisk, Winson Han,
  Roozbeh Mottaghi, Luke Zettlemoyer, and Dieter Fox.
\newblock {ALFRED: A Benchmark for Interpreting Grounded Instructions for
  Everyday Tasks}.
\newblock In \emph{Proceedings of the IEEE/CVF Conference on Computer Vision
  and Pattern Recognition (CVPR)}, June 2020.

\bibitem[Shukor et~al.(2025)Shukor, Aubakirova, Capuano, Kooijmans, Palma,
  Zouitine, Aractingi, Pascal, Russi, Marafioti, Alibert, Cord, Wolf, and
  Cadene]{shukor2025smolvla}
Mustafa Shukor, Dana Aubakirova, Francesco Capuano, Pepijn Kooijmans, Steven
  Palma, Adil Zouitine, Michel Aractingi, Caroline Pascal, Martino Russi,
  Andres Marafioti, Simon Alibert, Matthieu Cord, Thomas Wolf, and Remi Cadene.
\newblock {SmolVLA: A Vision-Language-Action Model for Affordable and Efficient
  Robotics}, 2025.
\newblock URL \url{https://arxiv.org/abs/2506.01844}.

\bibitem[Song et~al.(2025)Song, Ou, Kong, Li, Neubig, and
  Yue]{song2025visualpuzzles}
Yueqi Song, Tianyue Ou, Yibo Kong, Zecheng Li, Graham Neubig, and Xiang Yue.
\newblock {VisualPuzzles: Decoupling Multimodal Reasoning Evaluation from
  Domain Knowledge}, 2025.
\newblock URL \url{https://arxiv.org/abs/2504.10342}.

\bibitem[Srivastava et~al.(2021)Srivastava, Li, Lingelbach, Martín-Martín,
  Xia, Vainio, Lian, Gokmen, Buch, Liu, Savarese, Gweon, Wu, and
  Fei-Fei]{srivastava2021behaviorbenchmarkeverydayhousehold}
Sanjana Srivastava, Chengshu Li, Michael Lingelbach, Roberto Martín-Martín,
  Fei Xia, Kent Vainio, Zheng Lian, Cem Gokmen, Shyamal Buch, C.~Karen Liu,
  Silvio Savarese, Hyowon Gweon, Jiajun Wu, and Li~Fei-Fei.
\newblock {BEHAVIOR: Benchmark for Everyday Household Activities in Virtual,
  Interactive, and Ecological Environments}, 2021.
\newblock URL \url{https://arxiv.org/abs/2108.03332}.

\bibitem[Stojanovski et~al.(2025)Stojanovski, Stanley, Sharratt, Jones,
  Adefioye, Kaddour, and Köpf]{stojanovski2025reasoning}
Zafir Stojanovski, Oliver Stanley, Joe Sharratt, Richard Jones, Abdulhakeem
  Adefioye, Jean Kaddour, and Andreas Köpf.
\newblock {REASONING GYM: Reasoning Environments for Reinforcement Learning
  with Verifiable Rewards}.
\newblock In \emph{Advances in Neural Information Processing Systems
  (NeurIPS)}, 2025.

\bibitem[Tang et~al.(2025)Tang, Sun, Zhao, Yang, Lin, Zhang, Hou, Lu, Liu, and
  Han]{tang2025vlash}
Jiaming Tang, Yufei Sun, Yilong Zhao, Shang Yang, Yujun Lin, Zhuoyang Zhang,
  James Hou, Yao Lu, Zhijian Liu, and Song Han.
\newblock {VLASH: Real-Time VLAs via Future-State-Aware Asynchronous
  Inference}, 2025.
\newblock URL \url{https://arxiv.org/abs/2512.01031}.

\bibitem[Teoh et~al.(2025)Teoh, Lin, Li, Liu, Sollomoni, Harel, and
  Dong]{teoh2025captchas}
Xiwen Teoh, Yun Lin, Siqi Li, Ruofan Liu, Avi Sollomoni, Yaniv Harel, and
  Jin~Song Dong.
\newblock {Are {CAPTCHAs} Still Bot-hard? Generalized Visual {CAPTCHA} Solving
  with Agentic Vision Language Model}.
\newblock In \emph{34th USENIX Security Symposium (USENIX Security 25)}, 2025.

\bibitem[Wang et~al.(2026{\natexlab{a}})Wang, Lu, Wang, Bai, Liu, Zhang, Wang,
  Hu, Xie, Bai, Liu, Shen, Lin, and Yu]{wang2026cuagym}
Bowen Wang, Dunjie Lu, Junli Wang, Tianyi Bai, Shixuan Liu, Zhipeng Zhang,
  Haiquan Wang, Hao Hu, Tianbao Xie, Shuai Bai, Dayiheng Liu, Que Shen, Junyang
  Lin, and Tao Yu.
\newblock {CUA-Gym: Scaling Verifiable Training Environments and Tasks for
  Computer-Use Agents}, 2026{\natexlab{a}}.
\newblock URL \url{https://arxiv.org/abs/2605.25624}.

\bibitem[Wang et~al.(2025{\natexlab{a}})Wang, Zhu, Zhou, Li, He, Wei, and
  Xiong]{wang2025cognition}
Junyu Wang, Changjia Zhu, Yuanbo Zhou, Lingyao Li, Xu~He, Mingkui Wei, and
  Junjie Xiong.
\newblock {COGNITION: From Evaluation to Defense against Multimodal LLM CAPTCHA
  Solvers}, 2025{\natexlab{a}}.
\newblock URL \url{https://arxiv.org/abs/2512.02318}.
\newblock USENIX Security 2026.

\bibitem[Wang et~al.(2025{\natexlab{b}})Wang, Yin, Zhang, Zhang, Wang, Wang,
  Zhang, Chandrasegaran, Liu, Krishna, Xie, Wu, Fei-Fei, and
  Li]{wang2025mindcube}
Qineng Wang, Baiqiao Yin, Pingyue Zhang, Jianshu Zhang, Kangrui Wang, Zihan
  Wang, Jieyu Zhang, Keshigeyan Chandrasegaran, Han Liu, Ranjay Krishna,
  Saining Xie, Jiajun Wu, Li~Fei-Fei, and Manling Li.
\newblock {MindCube: Spatial Mental Modeling from Limited Views},
  2025{\natexlab{b}}.
\newblock URL \url{https://arxiv.org/abs/2506.21458}.

\bibitem[Wang et~al.(2025{\natexlab{c}})Wang, Zhuang, and
  Wu]{wang2025lvlmplayground}
Xinyu Wang, Bohan Zhuang, and Qi~Wu.
\newblock {Are Large Vision Language Models Good Game Players?}
\newblock In \emph{International Conference on Learning Representations
  (ICLR)}, 2025{\natexlab{c}}.
\newblock arXiv:2503.02358.

\bibitem[Wang et~al.(2026{\natexlab{b}})Wang, Zhang, Ge, Lian, Fu, Dunlap,
  Goldberg, Wang, Stoica, Chan, Min, and Gonzalez]{wang2026visgym}
Zirui Wang, Junyi Zhang, Jiaxin Ge, Long Lian, Letian Fu, Lisa Dunlap, Ken
  Goldberg, XuDong Wang, Ion Stoica, David~M. Chan, Sewon Min, and Joseph~E.
  Gonzalez.
\newblock {VisGym: Diverse, Customizable, Scalable Environments for Multimodal
  Agents}, 2026{\natexlab{b}}.
\newblock URL \url{https://arxiv.org/abs/2601.16973}.

\bibitem[Wen et~al.(2026)Wen, Ye, Zhang, Yang, and Zhu]{wen2026realtime}
Yule Wen, Yixin Ye, Yanzhe Zhang, Diyi Yang, and Hao Zhu.
\newblock {Real-Time Reasoning Agents in Evolving Environments}.
\newblock In \emph{International Conference on Learning Representations
  (ICLR)}, 2026.
\newblock arXiv:2511.04898.

\bibitem[Wu et~al.(2023)Wu, Tang, Mitchell, and Li]{wu2023smartplay}
Yue Wu, Xuan Tang, Tom~M. Mitchell, and Yuanzhi Li.
\newblock {SmartPlay: A Benchmark for LLMs as Intelligent Agents}, 2023.

\bibitem[Wu et~al.(2025)Wu, Xue, Feng, Wang, and Song]{wu2025mcabench}
Zonglin Wu, Yule Xue, Yaoyao Feng, Xiaolong Wang, and Yiren Song.
\newblock {MCA-Bench: A Multimodal Benchmark for Evaluating CAPTCHA Robustness
  Against VLM-based Attacks}, 2025.
\newblock URL \url{https://arxiv.org/abs/2506.05982}.

\bibitem[Xie et~al.(2024)Xie, Zhang, Chen, Li, Zhao, Cao, Hua, Cheng, Shin,
  Lei, Liu, Xu, Zhou, Savarese, Xiong, Zhong, and
  Yu]{xie2024osworldbenchmarkingmultimodalagents}
Tianbao Xie, Danyang Zhang, Jixuan Chen, Xiaochuan Li, Siheng Zhao, Ruisheng
  Cao, Toh~Jing Hua, Zhoujun Cheng, Dongchan Shin, Fangyu Lei, Yitao Liu,
  Yiheng Xu, Shuyan Zhou, Silvio Savarese, Caiming Xiong, Victor Zhong, and Tao
  Yu.
\newblock {OSWorld: Benchmarking Multimodal Agents for Open-Ended Tasks in Real
  Computer Environments}, 2024.
\newblock URL \url{https://arxiv.org/abs/2404.07972}.

\bibitem[Xu et~al.(2026)Xu, Xu, Yi, Yuan, Guang, Long, Chen, Wu, Yu, and
  Wang]{xu2026vsbench}
Zelai Xu, Zhexuan Xu, Xiangmin Yi, Huining Yuan, Mo~Guang, Kaiwen Long, Xinlei
  Chen, Yi~Wu, Chao Yu, and Yu~Wang.
\newblock {{VS-Bench}: Evaluating {VLMs} for Strategic Abilities in Multi-Agent
  Environments}.
\newblock In \emph{IEEE/CVF Conference on Computer Vision and Pattern
  Recognition (CVPR)}, 2026.
\newblock arXiv:2506.02387.

\bibitem[Yang et~al.(2025{\natexlab{a}})Yang, Yang, Gupta, Han, Fei-Fei, and
  Xie]{Yang_2025_CVPR}
Jihan Yang, Shusheng Yang, Anjali~W. Gupta, Rilyn Han, Li~Fei-Fei, and Saining
  Xie.
\newblock {Thinking in Space: How Multimodal Large Language Models See,
  Remember, and Recall Spaces}.
\newblock In \emph{Proceedings of the IEEE/CVF Conference on Computer Vision
  and Pattern Recognition (CVPR)}, pp.\  10632--10643, June 2025{\natexlab{a}}.

\bibitem[Yang et~al.(2025{\natexlab{b}})Yang, Lieret, Jimenez, Wettig,
  Khandpur, Zhang, Hui, Press, Schmidt, and Yang]{yang2025swesmith}
John Yang, Kilian Lieret, Carlos~E. Jimenez, Alexander Wettig, Kabir Khandpur,
  Yanzhe Zhang, Binyuan Hui, Ofir Press, Ludwig Schmidt, and Diyi Yang.
\newblock {SWE-smith: Scaling Data for Software Engineering Agents},
  2025{\natexlab{b}}.
\newblock URL \url{https://arxiv.org/abs/2504.21798}.

\bibitem[Yang et~al.(2025{\natexlab{c}})Yang, Chen, Zhang, Zhao, Qian, Wang,
  Wang, Koripella, Movahedi, Li, Ji, Zhang, and
  Zhang]{yang2025embodiedbenchcomprehensivebenchmarkingmultimodal}
Rui Yang, Hanyang Chen, Junyu Zhang, Mark Zhao, Cheng Qian, Kangrui Wang,
  Qineng Wang, Teja~Venkat Koripella, Marziyeh Movahedi, Manling Li, Heng Ji,
  Huan Zhang, and Tong Zhang.
\newblock {EmbodiedBench: Comprehensive Benchmarking Multi-modal Large Language
  Models for Vision-Driven Embodied Agents}, 2025{\natexlab{c}}.
\newblock URL \url{https://arxiv.org/abs/2502.09560}.

\bibitem[Yang et~al.(2024)Yang, Sun, Weihs, VanderBilt, Herrasti, Han, Wu,
  Haber, Krishna, Liu, Callison-Burch, Yatskar, Kembhavi, and
  Clark]{yang2024holodeck}
Yue Yang, Fan-Yun Sun, Luca Weihs, Eli VanderBilt, Alvaro Herrasti, Winson Han,
  Jiajun Wu, Nick Haber, Ranjay Krishna, Lingjie Liu, Chris Callison-Burch,
  Mark Yatskar, Aniruddha Kembhavi, and Christopher Clark.
\newblock {Holodeck: Language Guided Generation of 3D Embodied AI
  Environments}.
\newblock In \emph{Proceedings of the IEEE/CVF Conference on Computer Vision
  and Pattern Recognition (CVPR)}, 2024.

\bibitem[Ying et~al.(2026)Ying, Truong, Sharma, Zhao, Cloos, Allen, Griffiths,
  Collins, Hernández-Orallo, Isola, Gershman, and
  Tenenbaum]{ying2026aigamestorescalableopenended}
Lance Ying, Ryan Truong, Prafull Sharma, Kaiya~Ivy Zhao, Nathan Cloos,
  Kelsey~R. Allen, Thomas~L. Griffiths, Katherine~M. Collins, José
  Hernández-Orallo, Phillip Isola, Samuel~J. Gershman, and Joshua~B.
  Tenenbaum.
\newblock {AI Gamestore: Scalable, Open-Ended Evaluation of Machine General
  Intelligence with Human Games}, 2026.
\newblock URL \url{https://arxiv.org/abs/2602.17594}.

\bibitem[Yuan et~al.(2026)Yuan, Zhou, Xiong, et~al.]{yuan2026osworld2}
Mengqi Yuan, Zilong Zhou, Xinzhuang Xiong, et~al.
\newblock {OSWorld 2.0: Benchmarking Computer Use Agents on Long-Horizon
  Real-World Tasks}, 2026.

\bibitem[Yue et~al.(2024)Yue, Ni, Zhang, Zheng, Liu, Zhang, Stevens, Jiang,
  Ren, Sun, Wei, Yu, Yuan, Sun, Yin, Zheng, Yang, Liu, Huang, Sun, Su, and
  Chen]{Yue_2024_CVPR}
Xiang Yue, Yuansheng Ni, Kai Zhang, Tianyu Zheng, Ruoqi Liu, Ge~Zhang, Samuel
  Stevens, Dongfu Jiang, Weiming Ren, Yuxuan Sun, Cong Wei, Botao Yu, Ruibin
  Yuan, Renliang Sun, Ming Yin, Boyuan Zheng, Zhenzhu Yang, Yibo Liu, Wenhao
  Huang, Huan Sun, Yu~Su, and Wenhu Chen.
\newblock {MMMU: A Massive Multi-discipline Multimodal Understanding and
  Reasoning Benchmark for Expert AGI}.
\newblock In \emph{Proceedings of the IEEE/CVF Conference on Computer Vision
  and Pattern Recognition (CVPR)}, pp.\  9556--9567, June 2024.

\bibitem[Yue et~al.(2025)Yue, Zheng, Ni, Wang, Zhang, Tong, Sun, Yu, Zhang,
  Sun, Su, Chen, and Neubig]{yue-etal-2025-mmmu}
Xiang Yue, Tianyu Zheng, Yuansheng Ni, Yubo Wang, Kai Zhang, Shengbang Tong,
  Yuxuan Sun, Botao Yu, Ge~Zhang, Huan Sun, Yu~Su, Wenhu Chen, and Graham
  Neubig.
\newblock {{MMMU}-Pro: A More Robust Multi-discipline Multimodal Understanding
  Benchmark}.
\newblock In Wanxiang Che, Joyce Nabende, Ekaterina Shutova, and Mohammad~Taher
  Pilehvar (eds.), \emph{Proceedings of the 63rd Annual Meeting of the
  Association for Computational Linguistics (Volume 1: Long Papers)}, pp.\
  15134--15186, Vienna, Austria, July 2025. Association for Computational
  Linguistics.
\newblock ISBN 979-8-89176-251-0.
\newblock \doi{10.18653/v1/2025.acl-long.736}.
\newblock URL \url{https://aclanthology.org/2025.acl-long.736/}.

\bibitem[Zhang et~al.(2026)Zhang, Griffiths, Narasimhan, and
  Press]{zhang2026videogamebenchvisionlanguagemodelscomplete}
Alex~L. Zhang, Thomas~L. Griffiths, Karthik~R. Narasimhan, and Ofir Press.
\newblock {VideoGameBench: Can Vision-Language Models complete popular video
  games?}, 2026.
\newblock URL \url{https://arxiv.org/abs/2505.18134}.

\bibitem[Zhang et~al.(2024)Zhang, Cai, and Lee]{zhang2024vinoground}
Jianrui Zhang, Mu~Cai, and Yong~Jae Lee.
\newblock {Vinoground: Scrutinizing LMMs over Dense Temporal Reasoning with
  Short Videos}, 2024.
\newblock URL \url{https://arxiv.org/abs/2410.02763}.

\bibitem[Zhang et~al.(2025)Zhang, Peng, Kong, Yang, Wu, Yu, Xiang, Ruan, Wang,
  Song, Liu, Tang, Liu, Wu, and Luo]{zhang2025autoenv}
Jiayi Zhang, Yiran Peng, Fanqi Kong, Cheng Yang, Yifan Wu, Zhaoyang Yu, Jinyu
  Xiang, Jianhao Ruan, Jinlin Wang, Maojia Song, HongZhang Liu, Xiangru Tang,
  Bang Liu, Chenglin Wu, and Yuyu Luo.
\newblock {AutoEnv: Automated Environments for Measuring Cross-Environment
  Agent Learning}, 2025.
\newblock URL \url{https://arxiv.org/abs/2511.19304}.

\bibitem[Zhao et~al.(2025)Zhao, Yang, Yu, Gao, and Shou]{zhao2025worldgui}
Henry~Hengyuan Zhao, Kaiming Yang, Wendi Yu, Difei Gao, and Mike~Zheng Shou.
\newblock {{WorldGUI}: An Interactive Benchmark for Desktop {GUI} Automation
  from Any Starting Point}, 2025.

\bibitem[Zheng et~al.(2025{\natexlab{a}})Zheng, Li, Yang, Yu, Wang, Yan, Yao,
  and Wang]{zheng2025vmage}
Xiangxi Zheng, Linjie Li, Zhengyuan Yang, Ping Yu, Alex~Jinpeng Wang, Rui Yan,
  Yuan Yao, and Lijuan Wang.
\newblock {{V-MAGE}: A Game Evaluation Framework for Assessing Vision-Centric
  Capabilities in Multimodal Large Language Models}, 2025{\natexlab{a}}.

\bibitem[Zheng et~al.(2025{\natexlab{b}})Zheng, Lin, He, Wang, Zheng, and
  Liang]{zheng2025mcu}
Xinyue Zheng, Haowei Lin, Kaichen He, Zihao Wang, Zilong Zheng, and Yitao
  Liang.
\newblock {{MCU}: An Evaluation Framework for Open-Ended Game Agents}.
\newblock In \emph{International Conference on Machine Learning (ICML)},
  2025{\natexlab{b}}.
\newblock arXiv:2310.08367.

\bibitem[Zhou et~al.(2024)Zhou, Xu, Zhu, Zhou, Lo, Sridhar, Cheng, Ou, Bisk,
  Fried, Alon, and Neubig]{zhou2024webarena}
Shuyan Zhou, Frank~F. Xu, Hao Zhu, Xuhui Zhou, Robert Lo, Abishek Sridhar,
  Xianyi Cheng, Tianyue Ou, Yonatan Bisk, Daniel Fried, Uri Alon, and Graham
  Neubig.
\newblock {{WebArena}: A Realistic Web Environment for Building Autonomous
  Agents}.
\newblock In \emph{International Conference on Learning Representations
  (ICLR)}, 2024.
\newblock arXiv:2307.13854.

\end{thebibliography}
\bibliographystyle{iclr2027_conference}

\clearpage
\appendix

\section*{Appendix Contents}
\startcontents[appendix]
\printcontents[appendix]{}{1}{\setcounter{tocdepth}{2}}

\section{Limitations}
\label{app:limitations}

We focus on evaluating existing agents, although our programmatic task generators and automatic verifiers also provide a basis for training. Studying whether training on these minigames improves real-time interaction is an important direction for future work.

Our minigames capture specific demands of real-time environments to enable controlled comparisons across capabilities, difficulty, and interaction. Whether improvements on these tasks transfer to complete games and physical interfaces remains an open question.

Finally, \bench{} covers a selected set of capabilities and task mechanics. Our creation--audit framework provides a way to extend this coverage with additional minigames and capability combinations.

\section{Minigame Construction}
\label{app:construction}

This appendix details the construction procedure summarized in Section~\ref{sec:methodology}. All agents in the construction pipeline use GPT-6 Astra.

\subsection{Creator and Audit Agents}
\label{app:creator-audit}

The creator agent receives one selected minigame together with its survey records and source links. It also inspects previously approved minigames as working examples of the environment structure. The creator implements the environment specification, task, generator, browser interface, server grader, exported verifier, oracle solver, difficulty and interaction controls, real-time settings, and tests. It may create or edit only the files of its own minigame and cannot modify the shared runtime, evaluator, or clock. The creator runs all ten difficulty and interaction configurations in the browser in the live and paused settings and checks generation, grading, result export, verification, failure, and retry. It records screenshots, recordings, and command outputs as evidence, together with a solution video in which the oracle solver completes the task in the browser.

A fresh audit agent then inspects the environment and its evidence without modifying any files. It checks the following items and ends its report with a verdict of \texttt{PASS} or \texttt{REVISION\_REQUIRED}.
\begin{enumerate}
    \item[(a)] \textbf{Difficulty baseline.} The generated task is assigned to the appropriate level by comparison with approved reference minigames, without relying on the environment name or the creator's description.
    \item[(b)] \textbf{Difficulty variations.} Levels 1 to 5 increase in what the agent must perceive, decide, remember, or control. Adding repetitive steps is not sufficient.
    \item[(c)] \textbf{Interaction.} For the same seed and difficulty, simplified and full interaction use the same minigame, information, goal, timing, physics, tolerances, and success condition. Full interaction requires the intended mouse or keyboard operations on the minigame.
    \item[(d)] \textbf{Real time.} In live execution, the environment continues while the model produces an action. In paused execution, it remains frozen until the action arrives, and the action then executes with the intended behavior. The real-time configurations satisfy the definition in Appendix~\ref{app:real-time-definition}, and no mechanism lets the agent avoid acting in real time, such as a brake that halts moving parts or repeating motion that permits a precomputed solution.
    \item[(e)] \textbf{End-to-end behavior.} All configurations load, remain playable, produce the correct grading result, export the result, handle failure, and allow another attempt.
    \item[(f)] \textbf{Evidence.} The audit agent ignores comments and written claims that the running environment does not support. Screenshots, recordings, frame sequences, exported results, and browser behavior serve as evidence, and missing visual evidence is a severe issue.
\end{enumerate}
The audit agent also checks that the creator changed only files within its permitted scope. The creator revises the implementation in response to the report, and a new audit agent repeats the audit.

\subsection{Difficulty Levels}
\label{app:difficulty}

The five difficulty levels, from very easy (level 1) to very hard (level 5), are defined in the same way for every minigame in \bench{}. A level describes the exact configuration the agent receives, and the original configuration of a minigame can belong to any level. A difficulty-review agent assigns levels by comparison with approved reference minigames, including Isometric Voxel Extraction Mine (level 1), Cursor-Controlled Constellation Hunt (level 2), Gyroscopic Tilt Board (level 3), Blind Dice Courier (level 4), and Exact-Change Candy Cascade (level 5).

To construct the other levels, the creator changes parameters that alter the problem the agent must solve. Depending on the minigame, these include the information available to the agent, visual ambiguity, the number of variables handled together, dependencies between decisions, planning depth, motion speed and timing windows, action precision, and recovery options. More steps, rounds, or waiting do not by themselves make a configuration harder. For every pair of adjacent levels, the creator must state what changes in the agent's decision or control problem. Table~\ref{tab:difficulty-example} shows the five levels of Rotating On-Screen Keyboard.

\begin{table}[t]
    \centering
    \caption{\textbf{Difficulty levels of Rotating On-Screen Keyboard.} The keyboard rotates in three dimensions with irregularly timed motion that never repeats within an attempt.}
    \label{tab:difficulty-example}
    \small
    \begin{tabular}{@{}clll@{}}
        \toprule
        \textbf{Level} & \textbf{Code length} & \textbf{Keys} & \textbf{Keyboard motion} \\
        \midrule
        1 & 3 & 12 large keys & Gentle, begins after the second character \\
        2 & 4 & 20 keys & Begins after the first character \\
        3 & 5 & 26 keys & Faster, begins after the first character \\
        4 & 5 & 32 keys & Faster, changes direction more often \\
        5 & 8 & 32 smaller keys & Fast, sharply irregular, starts immediately \\
        \bottomrule
    \end{tabular}
\end{table}

\subsection{Interaction Modes}
\label{app:interaction}

Section~\ref{sec:controlled-variations} defines the two interaction modes. For a fixed seed and difficulty, both modes generate the same world, information, goal, and action effects. The grader rejects actions submitted through the controls of the other mode. Table~\ref{tab:interaction-examples} shows examples.

\begin{table}[t]
    \centering
    \caption{\textbf{Examples of simplified and full interaction.}}
    \label{tab:interaction-examples}
    \small
    \begin{tabular}{@{}p{3.1cm}p{4.6cm}p{5.2cm}@{}}
        \toprule
        \textbf{Minigame} & \textbf{Simplified} & \textbf{Full} \\
        \midrule
        Rotating On-Screen Keyboard & Type the code on the physical keyboard. & Click each on-screen key while it moves. \\
        Clockwork Clutch Safe & Click a labeled release or engage button beneath each seal. & Drag each seal's clutch lever between its coupled and free positions. \\
        Parallel Grillmaster & Select a food, then click buttons to start cooking and move it to the tray. & Drag each food from the order to the grill and then to the serving tray. \\
        LIDAR Blacksite & Hold labeled movement and turning buttons, and use buttons to scan, pick up, and verify. & Hold movement keys, drag the view to turn, click the scene to scan or pick up the beacon, and walk through the exit gate. \\
        \bottomrule
    \end{tabular}
\end{table}

\subsection{Real-Time Revisions}
\label{app:real-time-revisions}

We revised minigames whose earlier versions did not require real-time interaction or allowed an agent to avoid it. For 64 minigames, a creator agent developed a new real-time minigame from the same source, as described in Section~\ref{sec:creator-auditor}. In 35 of these, the conversion removed the need for exploration, and a further revision restored it while keeping the goal, mechanic, and real-time execution. We also revised 23 minigames that already had real-time configurations, for instance to remove a brake that halts moving parts or to replace repeating motion that permits a precomputed solution. The remaining 13 minigames keep their original implementation, and \bench{} uses the revised versions of all others.

\subsection{Non-Real-Time Counterparts}
\label{app:counterparts}

Section~\ref{sec:counterparts} describes how we construct a non-real-time counterpart for each real-time minigame. The counterpart keeps the generator, goal, controls, difficulty parameters, available information, and grader of its real-time minigame, and changes only how the environment evolves over time. For example, in Rotating On-Screen Keyboard, the keyboard keeps its initial layout while the agent decides. In Clockwork Clutch Safe, each drive segment holds until the agent operates a clutch, and in Parallel Grillmaster, each food settles at its golden point once cooking starts. An independent audit agent compares the two versions across all ten configurations and returns any difference other than timing to the creator for repair. For the 35 minigames revised to restore exploration, we rebuilt the counterparts from the revised versions.

\section{Minigame List}
\label{app:minigame-list}

Table~\ref{tab:minigame-list} lists the 100 real-time minigames in \bench{} and the capabilities that each minigame and its non-real-time counterpart require.

\begin{table}[p]
    \centering
    \caption{\textbf{Minigames in \bench{}.} Capability labels: 3D visual understanding (3D), temporal reasoning (T), reasoning and planning (R), and exploration and interface understanding (E). A capability is marked when it is required at some difficulty level. $\bullet$: required by the real-time minigame and its non-real-time counterpart. $\circ$: required only by the real-time minigame. $\diamond$: required only by the counterpart.}
    \label{tab:minigame-list}
    \scriptsize
    \setlength{\tabcolsep}{3pt}
    \begin{tabular}{@{}lcccc@{\hspace{12pt}}lcccc@{}}
        \toprule
        \textbf{Minigame} & \textbf{3D} & \textbf{T} & \textbf{R} & \textbf{E} & \textbf{Minigame} & \textbf{3D} & \textbf{T} & \textbf{R} & \textbf{E} \\
        \midrule
        After Hours at the Reliquary &  & $\circ$ & $\bullet$ & $\bullet$ & Impossible Ecology &  & $\bullet$ & $\bullet$ & $\bullet$ \\
        Anthill Front &  & $\bullet$ & $\bullet$ & $\bullet$ & Impossible Panorama & $\bullet$ & $\bullet$ & $\bullet$ & $\bullet$ \\
        Apothecary Dead Reckoning &  & $\bullet$ & $\bullet$ & $\bullet$ & Isometric Voxel Extraction Mine & $\bullet$ & $\circ$ & $\bullet$ & $\bullet$ \\
        Ballast Lantern &  & $\bullet$ & $\bullet$ &  & Lampwrights Program & $\bullet$ & $\circ$ & $\bullet$ &  \\
        Bandaged Rose Window &  & $\bullet$ & $\bullet$ & $\bullet$ & Lantern Loft & $\bullet$ & $\bullet$ & $\bullet$ &  \\
        Blind Corridor Oscilloscope &  & $\bullet$ & $\bullet$ & $\bullet$ & Lanternfin Dive & $\bullet$ & $\bullet$ & $\bullet$ &  \\
        Blind Dice Courier & $\bullet$ & $\bullet$ & $\bullet$ &  & Lanternwing Roundup & $\bullet$ & $\bullet$ & $\bullet$ &  \\
        Bomb Manual From Hell &  & $\circ$ & $\bullet$ &  & Letter Rapids &  & $\bullet$ & $\bullet$ &  \\
        Branch Repair & $\bullet$ & $\bullet$ & $\bullet$ & $\bullet$ & LIDAR Blacksite &  & $\bullet$ & $\bullet$ & $\bullet$ \\
        Bureaucratic Signature Trap &  & $\bullet$ & $\bullet$ & $\bullet$ & Live Control-Flow Wiring Lab &  & $\bullet$ & $\bullet$ & $\bullet$ \\
        Cell Gatekeeper &  & $\bullet$ & $\bullet$ &  & Living Scaffold &  & $\bullet$ & $\bullet$ & $\diamond$ \\
        Charter of the Nine Cantons &  & $\bullet$ & $\bullet$ & $\bullet$ & Magnetic-Stripe Purgatory &  & $\bullet$ & $\bullet$ & $\bullet$ \\
        Circle Limit Twist &  & $\circ$ & $\bullet$ & $\bullet$ & Motion-Only Ghost Jigsaw &  & $\bullet$ &  &  \\
        Clockbeat Catacomb &  & $\bullet$ & $\bullet$ &  & Museum of Lost Gestures &  & $\bullet$ & $\bullet$ &  \\
        Clockwork Clutch Safe &  & $\circ$ & $\bullet$ &  & Occlusion Shell Swindle &  & $\bullet$ & $\bullet$ & $\bullet$ \\
        Clockwork Doppelgänger Customs &  & $\bullet$ & $\bullet$ & $\bullet$ & Parallax / Inertial Jigsaw Alignment &  & $\bullet$ & $\bullet$ & $\bullet$ \\
        Cloudpost Circuit & $\bullet$ & $\bullet$ & $\bullet$ &  & Parallax Orchard & $\bullet$ & $\bullet$ & $\bullet$ & $\bullet$ \\
        Cloudstep Caddie & $\bullet$ & $\bullet$ & $\bullet$ &  & Parallel Grillmaster &  & $\circ$ & $\bullet$ &  \\
        Collision Chimes &  & $\circ$ & $\bullet$ &  & Pearl Lattice & $\bullet$ & $\bullet$ & $\bullet$ &  \\
        Concertina Courier &  & $\bullet$ & $\bullet$ &  & Photograph Eats the Room & $\circ$ & $\bullet$ & $\bullet$ & $\circ$ \\
        Consent Gauntlet &  & $\circ$ & $\bullet$ &  & Pocket Locksmith & $\bullet$ & $\bullet$ & $\bullet$ & $\bullet$ \\
        Consequences Boss &  & $\bullet$ & $\bullet$ &  & Polarized Palimpsest &  & $\bullet$ &  & $\bullet$ \\
        Coordinates by Another Name &  & $\bullet$ & $\bullet$ & $\bullet$ & Polycube Parcel & $\bullet$ & $\bullet$ & $\bullet$ & $\circ$ \\
        Courtesy Junction &  & $\bullet$ & $\bullet$ &  & Polyrhythm Customs &  & $\bullet$ & $\bullet$ &  \\
        Crackglaze Crossing &  & $\bullet$ & $\bullet$ & $\diamond$ & Popup Exorcist &  & $\bullet$ & $\bullet$ & $\bullet$ \\
        CRAFTCHA: Alchemy Bench &  & $\bullet$ & $\bullet$ &  & Rayglass Vault &  & $\bullet$ & $\bullet$ & $\bullet$ \\
        Crash-Deadline Hovercar &  & $\bullet$ & $\bullet$ &  & Recursive Dollhouse Smuggling &  & $\bullet$ & $\bullet$ &  \\
        Crater Walker &  & $\bullet$ & $\bullet$ & $\bullet$ & Reflow Vitrine &  & $\bullet$ & $\bullet$ & $\bullet$ \\
        Cursor-Controlled Constellation Hunt &  & $\bullet$ & $\bullet$ & $\bullet$ & Reload Interruption &  & $\bullet$ & $\bullet$ &  \\
        Dead Man's Switch &  & $\bullet$ & $\bullet$ &  & Reveal to Identify &  & $\bullet$ & $\bullet$ & $\bullet$ \\
        Downsky Causeway & $\bullet$ & $\bullet$ & $\bullet$ &  & Ribbon Switchboard &  & $\bullet$ & $\bullet$ & $\bullet$ \\
        Dual-Projection Sculpture Rig & $\bullet$ & $\circ$ & $\bullet$ & $\bullet$ & Rising Causeway & $\bullet$ & $\bullet$ & $\bullet$ & $\diamond$ \\
        Elastic Membrane Sorter &  & $\bullet$ & $\bullet$ &  & Rorschach / Subjective Prompt With A Fixed Rubric &  & $\bullet$ & $\bullet$ & $\bullet$ \\
        Elbow Engine &  & $\bullet$ & $\bullet$ &  & Rotate The Wrong Thing Upright & $\bullet$ & $\bullet$ & $\bullet$ & $\bullet$ \\
        Ember Anvil & $\circ$ & $\circ$ & $\bullet$ & $\bullet$ & Rotating On-Screen Keyboard & $\circ$ & $\circ$ &  &  \\
        Facet Lantern & $\bullet$ & $\bullet$ & $\bullet$ & $\bullet$ & Shadow Crime Lab &  & $\bullet$ & $\bullet$ & $\bullet$ \\
        Firewatch Fold &  & $\bullet$ & $\bullet$ & $\bullet$ & Slot-Reel Character Capture &  & $\circ$ &  &  \\
        First Change Memory &  & $\bullet$ & $\bullet$ & $\bullet$ & Surveyors Toybox & $\bullet$ & $\bullet$ & $\bullet$ &  \\
        Five-Second Rule &  & $\bullet$ & $\bullet$ &  & The Flat Prisoner & $\bullet$ & $\bullet$ & $\bullet$ & $\bullet$ \\
        Five-System Verification Reactor &  & $\bullet$ & $\bullet$ &  & The Last Carbon Isles &  & $\bullet$ & $\bullet$ &  \\
        Flip-Gate Cascade &  & $\bullet$ & $\bullet$ & $\bullet$ & The Unlabeled Drawer &  & $\circ$ & $\bullet$ & $\bullet$ \\
        Fluke Census &  & $\bullet$ & $\bullet$ & $\bullet$ & The Unwatched Wing &  & $\bullet$ & $\bullet$ & $\bullet$ \\
        Forced-Perspective Moving Day & $\bullet$ & $\circ$ & $\bullet$ &  & Three-Camera Claw Machine & $\bullet$ & $\bullet$ & $\bullet$ &  \\
        Four-Pane Pilgrimage &  & $\bullet$ & $\bullet$ & $\bullet$ & Tiny FPS Customs & $\bullet$ & $\circ$ & $\bullet$ & $\bullet$ \\
        Four-Tab Robot Handshake &  & $\bullet$ & $\bullet$ & $\diamond$ & Tomographic Baggage Surgery & $\bullet$ & $\bullet$ & $\bullet$ & $\bullet$ \\
        Funeral Ritual &  & $\bullet$ & $\bullet$ & $\bullet$ & Top-Face Dice Arithmetic & $\bullet$ & $\bullet$ & $\bullet$ &  \\
        Gyroscopic Tilt Board &  & $\bullet$ & $\bullet$ &  & Two-Lamp Dyeworks &  & $\bullet$ & $\bullet$ & $\bullet$ \\
        Hearthlift Courier & $\bullet$ & $\bullet$ & $\bullet$ & $\bullet$ & Wonky Text Under Hostile Rendering &  & $\bullet$ & $\bullet$ & $\bullet$ \\
        Hologram Silhouette Foundry & $\bullet$ & $\bullet$ & $\bullet$ &  & Wrong Number &  & $\circ$ & $\bullet$ & $\circ$ \\
        Horizon Relay & $\diamond$ & $\circ$ & $\bullet$ &  & Zero-G Cable Autopsy &  & $\bullet$ & $\bullet$ &  \\
        \bottomrule
    \end{tabular}
\end{table}

\section{Capability Annotation Criteria}
\label{app:capability-criteria}

\bench{} studies six capabilities. Review agents annotate the four capabilities below. The remaining two, interaction and real-time interaction, are varied through the interaction mode and the real-time setting. Appendix~\ref{app:real-time-definition} defines when a configuration requires real-time interaction.

The authors define the following criteria for capability annotation. Review agents apply them to the requirements of a solution using screenshots and visible mouse or keyboard controls.

\paragraph{Visual understanding.}
We classify visual understanding as 2D or 3D according to the spatial understanding required to solve the task. Decorative perspective alone is insufficient for a 3D label.

\paragraph{Temporal reasoning.}
A configuration requires this capability when every general screenshot-only solution must interpret change, motion, order, or duration, remember relevant earlier state that is no longer visible, or control when and for how long an action remains active. The temporal relationship must affect action choice or execution. A simple visible sequence, idle animation, or automatic transition to a static result does not establish this requirement.

\paragraph{Reasoning and planning.}
Solving requires inferring constraints or choosing actions whose consequences affect later decisions. The annotation identifies the relevant inference or action dependency. Basic visual recognition alone does not establish this requirement.

\paragraph{Exploration and interface understanding.}
The agent must interact to reveal relevant information or learn how the interface behaves before solving the task. Routine feedback, an automatic transition that supplies the next required information, or following explicit controls does not establish this requirement.

Review agents inspect the implementation and recorded behavior for each difficulty and interaction mode. Each annotation is supported by code or behavioral evidence.

\section{Real-Time Evaluation Protocol}
\label{app:real-time-protocol}

The shared runtime controls the execution schedule, and the task clock begins when the environment starts. In live execution, an observation request returns a current frame while the task clock continues. Paused execution advances the task through an observation window and returns a chronological sequence of frames. The clock then stops on the final frame until the action executes and resumes for the next observation window.

The observation window and number of frames are set per minigame and are the same for its non-real-time counterpart. The window ranges from 80 to 800\,ms with three to six frames.

\subsection{Action and Screenshot Latency}
\label{app:infrastructure}

\paragraph{Direct connection protocol.} Our action and screenshot path builds on a fast input and output path that we previously developed for a single agent controlling a single sandbox. That design assumes a one-to-one pairing between agent and sandbox. Gym-Anything~\citep{aggarwal2026gymanythingturnsoftwareagent}, in contrast, runs many sandboxes behind a central server that launches them and forwards requests to them, and forwarding every action and screenshot through this server adds latency. We therefore separate the two roles. The central server still allocates sandboxes and manages their lifecycle, but actions and screenshots bypass it. The agent's client runs on the same machine as its sandbox and keeps a persistent direct HTTP connection to an input service inside the sandbox, which executes each action and returns its acknowledgement over this connection. Screenshots are captured natively inside the sandbox as full-resolution JPEG images, encoded once, and returned without resizing or re-encoding. A single request can execute an action and return the subsequent screenshot.

\paragraph{Measurement.} We measure latency from immediately before the HTTP request until its complete response, including the image, has arrived. Model inference and subsequent image loading and decoding are outside this measurement.

In isolated timing tests with one client, we measured 150 sequential requests per condition after five warm-up requests. Median latencies were 39.2\,ms for Rotating On-Screen Keyboard, 42.5\,ms for an idle LIDAR Blacksite scene, and 46.8\,ms while issuing LIDAR scan clicks. The corresponding 95th percentiles were 52.1, 55.3, and 70.7\,ms.

\section{Evaluation Details}
\label{app:evaluation}

This appendix details the evaluation protocol summarized in Section~\ref{sec:benchmark-protocol}.

\subsection{Models and Harnesses}
\label{app:models}

We evaluate GPT-6 Astra with low and xHigh reasoning effort and GPT-6 Luna with xHigh reasoning effort through the Codex CLI. Each GPT model receives a single \texttt{computer} tool. We disable the shell, code execution, web search, browser, and built-in computer-use features. Claude Opus 5.5 with xHigh effort runs through Claude Code with all built-in tools disabled, and receives the same \texttt{computer} tool through an MCP server. Qwen3.5-9B and Qwen3.8-27B are served through the Tinker API with temperature 1.0, top-p 0.95, top-k 20, and at most 2,048 output tokens per response. They use their native computer-use function format with coordinates on a $1000\times1000$ grid, and their context contains the last 100 turns with at most 20 screenshots.

\subsection{Action Space and Prompt}
\label{app:prompt}

For the GPT models, each call to the \texttt{computer} tool contains an ordered list of mouse and keyboard actions and returns a screenshot after the actions complete. The actions include moving, clicking, dragging, scrolling, typing, pressing keys, holding and releasing keys and mouse buttons, and waiting. The model instructions and the text after each task description both instruct all models to use only the visible controls of the task webpage, without writing code or inspecting the page. Figure~\ref{lst:prompt} shows the complete prompt for Rotating On-Screen Keyboard in the live setting. In the paused setting, the paragraph following \texttt{Temporal mode} is replaced by the text in Figure~\ref{lst:paused}.

\begin{figure}[p]
\begin{Verbatim}[fontsize=\scriptsize,breaklines=true,breakanywhere=true,frame=single,framesep=6pt,breaksymbolleft={},breaksymbolright={}]
Solve the task using your own visual reasoning and the computer tool. Choose literal mouse/keyboard inputs or request a screenshot with actions=[]. Writing or executing code, scripts, loops, OCR, automated image processing, and background automation is prohibited. Use only visible controls of the task webpage. Do not open a terminal, external application, browser developer tools, console, inspector, source, DOM, or network panel. Do not navigate, reload, change URLs, inspect hidden state, or switch to unrelated tabs.

Use the computer tool with {"actions": [...]} for all observations and inputs. Every call returns the screenshot after its inputs complete. Start with actions=[].

Coordinates use the returned 1920x1080 screenshot at its original
resolution: x increases rightward and y downward from [0, 0]. Coordinates
are passed directly to the task display.

Native env.step input objects (use the fields needed for your action):
- Mouse point fields: move, left_click, right_click, middle_click, double_click, triple_click. Each takes [x,y],
  e.g. {"mouse":{"move":[100,200]}}.
- Mouse drag fields: left_click_drag, right_click_drag. Each takes [[x1,y1],[x2,y2]].
- "mouse": {"buttons": {...}} accepts boolean states: left_down, left_up, right_down, right_up, middle_down, middle_up.
- "mouse": {"scroll": n} sends n wheel notches, positive down, negative up.
  These are NOT pixels. To scroll at a point, move the pointer first.
- "keyboard" accepts text, keys, keys_down, keys_up. text is a string. keys,
  keys_down and keys_up accept a key name or a list of key names, e.g. "w",
  "shift", "ctrl", "left", "space", "Return". keys taps a chord. keys_down
  presses without releasing. keys_up releases. A separate wait after a tap
  does not hold the key. There is no duration field on keyboard or mouse.
- {"action":"wait", "time":0.2} waits in seconds ("seconds" is also accepted).
- {"action":"screenshot"} or [] observes without spending the step budget.
  A screenshot entry inside a batch does not yield an intermediate frame.

Keep distinct inputs in separate list entries when order matters. The runtime
processes mouse before keyboard within a combined object, so use keys_down,
mouse input, keys_up entries for modifier-held gestures. To draw a curved path,
send mouse down, a sequence of mouse moves (and waits if needed), then mouse up.
Inputs can stay held across requests and screenshots. Do not replace a held
input with repeated taps or type repeated letters to simulate movement.

Temporal mode: live
The task keeps running while you reason, deliver inputs, and wait.
Each response returns one instantaneous screenshot after the action batch
finishes. Held keys/buttons remain down across responses until you release them.
A batch containing down, wait, up can express a timed hold in live mode.
execute_at_s is not available in this mode.


You have at most 100 steps. Each non-observation request, including an ordered action list, consumes one step. Screenshot-only requests do not consume steps. Inputs in a batch execute in order. Release held inputs before finishing. Stop when done is true.

Task:
Confirm the shown code using the on-screen keyboard.

Solve the task using your own visual reasoning and the computer tool. Choose literal mouse/keyboard inputs or request a screenshot with actions=[]. Writing or executing code, scripts, loops, OCR, automated image processing, and background automation is prohibited. Use only visible controls of the task webpage. Do not open a terminal, external application, browser developer tools, console, inspector, source, DOM, or network panel. Do not navigate, reload, change URLs, inspect hidden state, or switch to unrelated tabs.

Your answer only counts once it is submitted: after completing the task, press the visible submit/confirm/verify control and check the UI acknowledges it before you declare the task finished.
\end{Verbatim}
\caption{\textbf{Prompt for Rotating On-Screen Keyboard in the live setting.}}
\label{lst:prompt}
\end{figure}

\begin{figure}[t]
\begin{Verbatim}[fontsize=\scriptsize,breaklines=true,breakanywhere=true,frame=single,framesep=6pt,breaksymbolleft={},breaksymbolright={}]
The task clock is paused while you reason and while inputs are delivered.
Each screenshot request or step response advances one configured observation
window, returning its frames in chronological order. The task then freezes on
the final frame, and the next input acts on that state. A wait sleeps in wall
time but does not advance the paused task clock. To hold a key or mouse button
while the task advances, send down in one request and up in a later request.
Additional screenshot requests advance additional windows while the input
stays held. Down, wait, up in a single batch does NOT advance paused task time
between down and up.
\end{Verbatim}
\caption{\textbf{Timing instructions in the paused setting.}}
\label{lst:paused}
\end{figure}

\subsection{Budgets and Outcomes}
\label{app:budgets}

Each attempt allows at most 100 steps. A step is one request that contains one or more actions in order. Requests that only return a screenshot do not consume steps. Each attempt has a time limit of six hours, except that attempts of Astra (xHigh) on the real-time tasks at $1\times$, $4\times$, and $16\times$ slowdown and in the paused setting have a limit of one hour. Attempts that reach the time limit count as failures. We exclude and rerun attempts that end with a provider or infrastructure error. The task verifier determines whether each attempt succeeds.

\subsection{Time and Cost}
\label{app:cost}

Time is the wall-clock duration of an attempt from the moment the minigame's clock starts to the end of its episode. It excludes the time spent setting up the environment. Cost is the API-equivalent cost of the recorded token usage,
\[
    \text{cost} = \frac{(n_{\text{input}} - n_{\text{cached}})\,p_{\text{input}} + n_{\text{cached}}\,p_{\text{cached}} + n_{\text{output}}\,p_{\text{output}}}{10^{6}},
\]
where reasoning tokens are counted once as output tokens. Table~\ref{tab:prices} lists the prices. For GPT-6 Astra and Luna, we use OpenAI API prices,\footnote{\url{https://developers.openai.com/api/docs/pricing}} and requests with more than 272K input tokens are charged twice the input and cached-input prices and 1.5 times the output price. For the Qwen models, we use OpenRouter prices as of October 2026.\footnote{\url{https://openrouter.ai}} OpenRouter lists no discounted price for cached input tokens of Qwen3.5-9B. We therefore charge its cached input tokens at the regular input price. For Claude Opus 5.5, we use the cost that Claude Code reports for each session, which applies Anthropic API prices to the recorded token usage.

\begin{table}[t]
    \centering
    \caption{\textbf{Prices used to compute cost}, in USD per million tokens.}
    \label{tab:prices}
    \small
    \begin{tabular}{@{}lrrr@{}}
        \toprule
        \textbf{Model} & \textbf{Input} & \textbf{Cached input} & \textbf{Output} \\
        \midrule
        Qwen3.5-9B & 0.17 & -- & 0.25 \\
        Qwen3.8-27B & 0.20 & 0.09 & 2.50 \\
        GPT-6 Luna & 0.10 & 0.01 & 0.50 \\
        GPT-6 Astra & 10.00 & 1.00 & 50.00 \\
        \bottomrule
    \end{tabular}
\end{table}

\subsection{Coding Protocol}
\label{app:coding-protocol}

In the coding experiment, the model runs in the Codex CLI with its standard tools inside an isolated sandbox. It receives an HTTP endpoint that accepts the same actions and returns screenshots, together with a short Python client. The model can write programs that process screenshots and issue actions through this endpoint. Access to the task machine, page source, and browser developer tools remains prohibited.

\subsection{Interface Rulebooks}
\label{app:rulebooks}

For the experiment in Section~\ref{sec:exploration-results}, each rulebook describes a minigame's controls, action effects, fixed world rules, feedback, submission, and recovery. It excludes generated answers, target coordinates, hidden state, and seed-specific layouts or solutions. The rulebook is appended to the task prompt, and all other settings remain unchanged.

\subsection{Human Baseline}
\label{app:human}

One of the authors played 20 tasks randomly sampled from the 100 real-time tasks at difficulty 5 with full interaction in the live setting, seeing each minigame for the first time. Each task used the same seed that the models received. The author solved 9 of the 20 tasks (45.0\%), with an average time of 132.7 seconds, or 2.2 minutes, per task.

\section{Examples of Model-Written Controllers}
\label{app:coding-examples}

We examine saved trajectories from Astra's coding evaluation at difficulty 5 with full interaction in the live setting. In Rotating On-Screen Keyboard, the model uses perspective transforms and image templates to locate requested keys, converts their positions to screen coordinates, and clicks them. In Wrong Number, it fits curves to waveform pixels from screenshots, estimates the phase error, and adjusts a slider. In Clockwork Clutch Safe, it tracks the marker on each gear, estimates the marker's angular speed from recent screenshots, and releases each clutch when the marker is predicted to reach its target angle. All three attempts pass the task verifier.

Image-based control also appears in failed attempts. In Gyroscopic Tilt Board, Astra detects the ball and direction indicator, estimates velocity, and moves the tilt control along a sequence of waypoints, but the attempt fails. These examples show how scripts can process observations and issue actions between model responses.

\FloatBarrier

\section{Real-Time Definition and Review Prompts}
\label{app:real-time-definition}

\subsection{Definition}

The real-time label is assigned to a fixed difficulty and interaction configuration. Let $H_t$ denote the visible interaction history through task time $t$, $I_t$ the agent's own prior actions and their task times, and $W_t^w$ the visible window from $t-w$ to $t$. Let $u$ denote the terminal task outcome, with less task-active continuation time after delivery of the next action used only to break ties between equal outcomes. Expectations use visible task information and average over future randomness. Private generator or page state is unavailable to the policy.

A grouped control mode $D$ contains every occurrence, across seeds, rounds, and attempts, of the same outcome-affecting control mode. Let $C_\Delta(D)$ contain the histories in $D$ from which success remains possible after a delay $\Delta$ with no new action. Only histories that become unsalvageable after this delay are excluded. Let $A^*_\Delta(H_t)$ be the set of actions maximizing expected $u$ when delivered at $t+\Delta$ and followed by optimal continuation. Let $H_t^\tau$ be the history after $\tau$ with no new action, and $Q(H_t,a)$ the expected $u$ when $a$ is the next action and is followed by optimal continuation.

A configuration satisfies the real-time definition if and only if there exist a grouped control mode $D$, a delay $\Delta>0$, and a bounded visual window $w$ such that $C_\Delta(D)$ is nonempty and all three conditions hold:
\begin{enumerate}
    \item[(i)] No policy using only the agent's own action history and task times selects an optimal delayed action for every viable history:
    \[
        \not\exists q:I_t\mapsto a
        \quad\text{such that}\quad
        \forall H_t\in C_\Delta(D),\;
        q(I_t)\in A^*_\Delta(H_t).
    \]
    \item[(ii)] A policy using that history and the recent visual window can select an optimal delayed action for every viable history:
    \[
        \exists g:(I_t,W_t^w)\mapsto a
        \quad\text{such that}\quad
        \forall H_t\in C_\Delta(D),\;
        g(I_t,W_t^w)\in A^*_\Delta(H_t).
    \]
    \item[(iii)] For some $H_t\in C_\Delta(D)$, action $a$, and $0<\tau<\Delta$, the environment evolves autonomously to $H_t^\tau$, success remains possible, and the action loses value or advantage over no action:
    \begin{align*}
        Q(H_t^\tau,a) &< Q(H_t,a)
        \qquad\text{or}\\
        Q(H_t^\tau,a)-Q(H_t^\tau,\mathrm{no\mbox{-}op})
        &< Q(H_t,a)-Q(H_t,\mathrm{no\mbox{-}op}).
    \end{align*}
\end{enumerate}

An unavailable or ineffective action has the value of no action. The strict loss in (iii) must arise from non-clock task-state evolution. Less remaining deadline, later completion alone, a fixed wait that enables an action, and a fixed animation whose result remains available do not satisfy this condition.

All three conditions must be witnessed by the same visible autonomous process and the same $D$, $\Delta$, and $w$. The evidence within $W_t^w$ must support prediction of the process's effect at $t+\Delta$ and determine the delayed-optimal action in (i) and (ii). The evidence can occur across frames or in an accumulated trace, chart, or trajectory within one screenshot. Unrelated animation and future randomness independent of the visible window cannot supply this evidence.

The witness delay $\Delta$ is chosen on the time scale of the task behavior. It begins within an active control opportunity and may extend beyond the point where that opportunity degrades or closes. It is independent of the current model provider's latency. The configured observation window and frame count bound $w$. A stationary policy over the visibly evidenced dynamic state is allowed.

\paragraph{Pre-run exception.}
For the configuration-level label, a configuration is classified as not real-time if at least 50\% of its generated instances admit a successful pre-run solution. After preparation, such a solution starts the autonomous outcome phase and sends no further outcome-affecting actions until that phase ends. Final submission or certification is administrative and is excluded from this action count. Reviewers establish this threshold from source evidence or a reproducible seed probe. If the threshold is unresolved and could change a positive mathematical classification, the reviewer returns \texttt{unresolved}. Failure of any mathematical condition already establishes a negative classification.

\subsection{Review Prompts}
\label{app:real-time-review-prompts}

The following are the prompt templates used to review real-time labels. The first template assigns one configuration to a reviewer. The second assigns the configurations of one minigame to a reviewer, with a separate judgment for each configuration. Both use the same definition and pre-run exception.

\subsubsection{One Configuration per Review}

\begin{Verbatim}
# Independent real-time classification prompt

Prompt version: `selected-100-pending-v1`

Reuses the historical real-time pilot prompt, adding the settled pre-run threshold and explicit report destination.

Frozen definition source: `docs/real-time.md`

Frozen definition SHA-256: `885cd87d0c59d5017b673a2b8696f251e31eedf0c8e388e32d47101625bcaf09`

Use the configuration and SOURCE_ROOT, CASE_INDEX, and OUTPUT_PATH values supplied by the launcher. Read source only from that frozen SOURCE_ROOT, not the current checkout. Paths below are relative to SOURCE_ROOT.

---

You are an independent reviewer classifying exactly one Gym-Anything-RT configuration.

## Configuration

- Public environment name: `{public_environment_name}`
- Environment ID: `{environment_id}`
- Difficulty: `{difficulty}`
- Interaction mode: `{interaction_mode}`

Decide whether this exact configuration is real-time under the frozen mathematical definition below.

Do not use an existing classification, capability annotation, audit result, result matrix, or another reviewer's answer. Do not infer the answer from the environment name or task description.

Read `AGENTS.md` and `docs/real-time.md` completely. Then read the complete relevant implementation:

1. the exact difficulty and interaction profile in `controls.json`;
2. the generator;
3. the visible browser implementation and relevant CSS;
4. the grader and verifier; and
5. the solver, when present.

Study a normal screenshot-only solution using visible task controls. Private state and scripted solvers are implementation evidence, not information available to the policy.

## Frozen mathematical definition

```text
H_t              visible interaction history through task time t
I_t              the agent's own prior actions and task times in H_t
W_t^w            the visible window from t-w through t
u                 terminal task outcome, with less task-active continuation
                  time after delivery of the next action used only to break
                  ties between equal outcomes
D                 one grouped control mode
C_Delta(D)        histories in D from which success remains possible after Delta
                  of no new action
A*_Delta(H_t)     actions maximising expected u when delivered at t+Delta,
                  followed by optimal continuation
H_t^tau           history tau later if the agent supplies no new action
Q(H_t, a)         expected u if a is the next action, followed by optimal continuation
```

Expectations use only visible task information and average over future randomness; private generator or page state is not available to the policy. For a fixed difficulty and interaction configuration, `D` contains every occurrence across seeds, rounds, and attempts of the same outcome-affecting control mode. It is not a hand-picked early interval. `C_Delta(D)` removes only histories whose outcome is already unsalvageable after the delay.

A fixed difficulty and interaction configuration of `E` is real-time iff there are a grouped control mode `D`, a delay `Delta > 0`, and a bounded visual window `w` such that `C_Delta(D)` is nonempty and:

```text
(i)    there is no q : I_t -> a
       with q(I_t) in A*_Delta(H_t) for every H_t in C_Delta(D)

(ii)   there is a g : (I_t, W_t^w) -> a
       with g(I_t, W_t^w) in A*_Delta(H_t)
       for every H_t in C_Delta(D)

(iii)  for some H_t in C_Delta(D), action a, and 0 < tau < Delta,
       the environment changes autonomously from H_t to H_t^tau,
       success remains possible from H_t^tau, and

       Q(H_t^tau, a) < Q(H_t, a)

       or

       Q(H_t^tau, a) - Q(H_t^tau, no-op)
         < Q(H_t, a) - Q(H_t, no-op).
```

An unavailable or ineffective action has the value of `no-op`. The strict loss in clause (iii) must be caused by non-clock task-state evolution, not merely by less remaining deadline or later completion. An action becoming enabled or more effective after a fixed wait does not satisfy the clause. A fixed animation that later enables controls or permanently archives its result is not an expiring opportunity.

The three clauses must be witnessed by the same phenomenon. The autonomous evolution in clause (iii) must be visibly evidenced within `W_t^w`, and that evidence must be what makes the delayed-optimal action depend on observation in clauses (i) and (ii). The window must support prediction of the evolution's effect at `t+Delta`; unrelated idle animation or future randomness independent of the window cannot supply clause (iii).

The evidence may appear across consecutive frames or as an accumulated visible trace, chart, or trajectory in one screenshot. A single screenshot is allowed when it visibly contains the relevant temporal evidence. This does not admit a static puzzle next to an unrelated animation.

`Delta` is a reported witness horizon on the time scale of the task behavior, not the current provider's response latency. It must begin inside a nontrivial active control opportunity and may extend across the point at which that opportunity gets worse or closes. The configured `observation_window_ms` and `frames_per_observation` bound `w`; they do not force every environment to need every captured frame.

The conjunction has these required consequences:

- Clause (i) rules out a fixed schedule or a policy driven only by the agent's own action clock.
- Clause (ii) requires a bounded recent visual window to be sufficient. This excludes unobservable randomness and information that must be recalled only from outside the window.
- Clause (iii) requires a useful action opportunity to get worse inside the delay because the task state changes. This excludes fixed waits, minimum-duration holds solvable from the agent's own clock, and information that remains actionable indefinitely.
- A stationary visual policy is allowed. It can map the currently evidenced changing state to an action without using absolute task time.
- All three clauses must use one common `D`, `Delta`, `w`, and visible autonomous process.

## Pre-run threshold

For the configuration-level catalog label, a configuration is Not real-time when
at least 50% of its generated instances admit a successful pre-run solution:
after preparation, the solution starts the autonomous outcome phase and sends
no more outcome-affecting actions until that phase ends. Final submission or
certification is administrative and does not count as an action during the run.

Record whether the threshold is met, with source-structural evidence or an actual reproducible seed probe. A failed existing solver does not prove that no pre-run solution exists. Do not invent a percentage. If the threshold cannot be determined and it could change a mathematical Yes into No, return unresolved. A configuration that already fails the mathematical definition is No regardless of an unresolved pre-run rate.

## Review procedure

1. State the controls available in this exact difficulty and interaction mode.
2. Describe every period in which the task changes without a new agent action.
3. State whether an outcome-affecting control remains available during each such period.
4. Enumerate the legitimate grouped control modes `D`.
5. Try to construct an observation-free policy `q` using only `I_t`.
6. Try to construct a bounded-window policy `g` using only `I_t` and `W_t^w`.
7. Check whether the same visible autonomous process causes an available action to become unavailable or strictly less effective within some `Delta`.
8. For `yes`, give one concrete common witness: `D`, `Delta`, `w`, the visible process, why `q` fails, why `g` succeeds, and the expiring or degrading action.
9. For `no`, identify the failed clause and explain why no legitimate choice of `D`, `Delta`, and `w` repairs it, or establish that the pre-run threshold overrides a mathematical Yes.
10. Return `unresolved` instead of guessing when the complete source does not determine the answer.

Return exactly one JSON object:

```json
{
  "case_index": "{case_index}",
  "public_environment_name": "{public_environment_name}",
  "environment_id": "{environment_id}",
  "difficulty": "{difficulty}",
  "interaction_mode": "{interaction_mode}",
  "label": "yes | no | unresolved",
  "controls_available": [],
  "autonomous_processes": [],
  "grouped_control_modes": [],
  "delta_ms": null,
  "window_ms": null,
  "clause_i": {"holds": true, "reason": ""},
  "clause_ii": {"holds": true, "reason": ""},
  "clause_iii": {"holds": true, "reason": ""},
  "common_visible_process": "",
  "pre_run": {
    "threshold": 0.5,
    "at_least_half": null,
    "rate": null,
    "basis": "source_structural | seeded_probe | unresolved | not_applicable",
    "reason": "",
    "seed_probe": null
  },
  "files_read_completely": [],
  "source_evidence": [
    {"path": "", "lines": "", "claim": ""}
  ],
  "uncertainties": []
}
```

Write the JSON object, with numeric case_index and difficulty, to the assigned OUTPUT_PATH using apply_patch. This one report is the only permitted persistent output. Do not edit puzzle sources, definitions, historical audits, or another report. Do not inspect other first-pass reviews, existing classification/capability files, evaluation results, or previous conversations. The audit manifest is parent-only; the launcher gives you your exact assignment.

You may run read-only source/math probes when needed, with PYTHONDONTWRITEBYTECODE=1 and without modifying source. For a seed probe, include seeds, numerator/denominator, and reproducible probe code in pre_run.seed_probe. No browser or desktop access, external websites/services, nested agents, or empirical CUA evaluation. A source-review report may be unresolved; do not guess to finish.

files_read_completely and source_evidence must use repository-relative paths. Read each complete implementation file needed by the assigned mechanic, including shared helpers actually used. If a shared file is long, read it in successive ranges to EOF; keyword searches do not substitute for full reads. Record concrete code evidence for the exact difficulty and interaction. Do not make changes to obtain a preferred label.
\end{Verbatim}

\subsubsection{Configurations Grouped by Minigame}

\begin{Verbatim}
# Per-game real-time classification prompt

Prompt version: `selected-100-per-game-v2`

Preserves the first-phase mathematical definition, pre-run threshold, and evidence requirements. Changes only the assignment unit: one reviewer reads one game once and judges its requested difficulty/interaction configurations together.

Frozen definition source: `docs/real-time.md`

Frozen definition SHA-256: `885cd87d0c59d5017b673a2b8696f251e31eedf0c8e388e32d47101625bcaf09`

Use the game assignment and SOURCE_ROOT, CASE_INDEX, CONFIGURATIONS, and OUTPUT_PATH values supplied by the launcher. Read source only from that frozen SOURCE_ROOT, not the current checkout. Paths below are relative to SOURCE_ROOT.

---

You are an independent reviewer classifying one Gym-Anything-RT game across its difficulty and interaction configurations in a single review.

## Assignment

The launcher supplies the exact public environment name, environment ID, game CASE_INDEX, and CONFIGURATIONS list. Each configuration has its own numeric case_index, difficulty, and interaction_mode. Normally the list contains all ten configurations: five difficulties crossed with Full and Simplified. A partially completed game may supply only its unfinished configurations. Read all ten active profiles together to understand their differences, but return judgments only for the supplied CONFIGURATIONS. Do not inspect or infer the already completed judgments.

Read each shared implementation file once, then apply the definition separately to every requested configuration. Explicitly check mode-specific inputs and level-dependent behavior; do not copy one answer across the game without checking those differences. These are multiple configuration judgments from one fresh game-level reviewer, not independent reviewer contexts per configuration.

Do not use an existing classification, capability annotation, audit result, result matrix, or another reviewer's answer. Do not infer the answer from the environment name or task description.

Read `AGENTS.md` and `docs/real-time.md` completely. Then read the complete relevant implementation:

1. the exact difficulty and interaction profile in `controls.json`;
2. the generator;
3. the visible browser implementation and relevant CSS;
4. the grader and verifier; and
5. the solver, when present.

Study a normal screenshot-only solution using visible task controls. Private state and scripted solvers are implementation evidence, not information available to the policy.

## Frozen mathematical definition

```text
H_t              visible interaction history through task time t
I_t              the agent's own prior actions and task times in H_t
W_t^w            the visible window from t-w through t
u                 terminal task outcome, with less task-active continuation
                  time after delivery of the next action used only to break
                  ties between equal outcomes
D                 one grouped control mode
C_Delta(D)        histories in D from which success remains possible after Delta
                  of no new action
A*_Delta(H_t)     actions maximising expected u when delivered at t+Delta,
                  followed by optimal continuation
H_t^tau           history tau later if the agent supplies no new action
Q(H_t, a)         expected u if a is the next action, followed by optimal continuation
```

Expectations use only visible task information and average over future randomness; private generator or page state is not available to the policy. For a fixed difficulty and interaction configuration, `D` contains every occurrence across seeds, rounds, and attempts of the same outcome-affecting control mode. It is not a hand-picked early interval. `C_Delta(D)` removes only histories whose outcome is already unsalvageable after the delay.

A fixed difficulty and interaction configuration of `E` is real-time iff there are a grouped control mode `D`, a delay `Delta > 0`, and a bounded visual window `w` such that `C_Delta(D)` is nonempty and:

```text
(i)    there is no q : I_t -> a
       with q(I_t) in A*_Delta(H_t) for every H_t in C_Delta(D)

(ii)   there is a g : (I_t, W_t^w) -> a
       with g(I_t, W_t^w) in A*_Delta(H_t)
       for every H_t in C_Delta(D)

(iii)  for some H_t in C_Delta(D), action a, and 0 < tau < Delta,
       the environment changes autonomously from H_t to H_t^tau,
       success remains possible from H_t^tau, and

       Q(H_t^tau, a) < Q(H_t, a)

       or

       Q(H_t^tau, a) - Q(H_t^tau, no-op)
         < Q(H_t, a) - Q(H_t, no-op).
```

An unavailable or ineffective action has the value of `no-op`. The strict loss in clause (iii) must be caused by non-clock task-state evolution, not merely by less remaining deadline or later completion. An action becoming enabled or more effective after a fixed wait does not satisfy the clause. A fixed animation that later enables controls or permanently archives its result is not an expiring opportunity.

The three clauses must be witnessed by the same phenomenon. The autonomous evolution in clause (iii) must be visibly evidenced within `W_t^w`, and that evidence must be what makes the delayed-optimal action depend on observation in clauses (i) and (ii). The window must support prediction of the evolution's effect at `t+Delta`; unrelated idle animation or future randomness independent of the window cannot supply clause (iii).

The evidence may appear across consecutive frames or as an accumulated visible trace, chart, or trajectory in one screenshot. A single screenshot is allowed when it visibly contains the relevant temporal evidence. This does not admit a static puzzle next to an unrelated animation.

`Delta` is a reported witness horizon on the time scale of the task behavior, not the current provider's response latency. It must begin inside a nontrivial active control opportunity and may extend across the point at which that opportunity gets worse or closes. The configured `observation_window_ms` and `frames_per_observation` bound `w`; they do not force every environment to need every captured frame.

The conjunction has these required consequences:

- Clause (i) rules out a fixed schedule or a policy driven only by the agent's own action clock.
- Clause (ii) requires a bounded recent visual window to be sufficient. This excludes unobservable randomness and information that must be recalled only from outside the window.
- Clause (iii) requires a useful action opportunity to get worse inside the delay because the task state changes. This excludes fixed waits, minimum-duration holds solvable from the agent's own clock, and information that remains actionable indefinitely.
- A stationary visual policy is allowed. It can map the currently evidenced changing state to an action without using absolute task time.
- All three clauses must use one common `D`, `Delta`, `w`, and visible autonomous process.

## Pre-run threshold

For the configuration-level catalog label, a configuration is Not real-time when
at least 50% of its generated instances admit a successful pre-run solution:
after preparation, the solution starts the autonomous outcome phase and sends
no more outcome-affecting actions until that phase ends. Final submission or
certification is administrative and does not count as an action during the run.

Record whether the threshold is met, with source-structural evidence or an actual reproducible seed probe. A failed existing solver does not prove that no pre-run solution exists. Do not invent a percentage. If the threshold cannot be determined and it could change a mathematical Yes into No, return unresolved. A configuration that already fails the mathematical definition is No regardless of an unresolved pre-run rate.

## Review procedure

1. State the controls available in this exact difficulty and interaction mode.
2. Describe every period in which the task changes without a new agent action.
3. State whether an outcome-affecting control remains available during each such period.
4. Enumerate the legitimate grouped control modes `D`.
5. Try to construct an observation-free policy `q` using only `I_t`.
6. Try to construct a bounded-window policy `g` using only `I_t` and `W_t^w`.
7. Check whether the same visible autonomous process causes an available action to become unavailable or strictly less effective within some `Delta`.
8. For `yes`, give one concrete common witness: `D`, `Delta`, `w`, the visible process, why `q` fails, why `g` succeeds, and the expiring or degrading action.
9. For `no`, identify the failed clause and explain why no legitimate choice of `D`, `Delta`, and `w` repairs it, or establish that the pre-run threshold overrides a mathematical Yes.
10. Return `unresolved` instead of guessing when the complete source does not determine the answer.

Return exactly one game-level JSON object with numeric game case_index, environment_id, public_environment_name, and configuration_reports. configuration_reports is an array containing exactly one object for each supplied configuration, with no duplicates or omitted or extra configurations. Each array member uses this schema (case_index is the configuration case_index, not the game case_index):

```json
{
  "case_index": "{case_index}",
  "public_environment_name": "{public_environment_name}",
  "environment_id": "{environment_id}",
  "difficulty": "{difficulty}",
  "interaction_mode": "{interaction_mode}",
  "label": "yes | no | unresolved",
  "controls_available": [],
  "autonomous_processes": [],
  "grouped_control_modes": [],
  "delta_ms": null,
  "window_ms": null,
  "clause_i": {"holds": true, "reason": ""},
  "clause_ii": {"holds": true, "reason": ""},
  "clause_iii": {"holds": true, "reason": ""},
  "common_visible_process": "",
  "pre_run": {
    "threshold": 0.5,
    "at_least_half": null,
    "rate": null,
    "basis": "source_structural | seeded_probe | unresolved | not_applicable",
    "reason": "",
    "seed_probe": null
  },
  "files_read_completely": [],
  "source_evidence": [
    {"path": "", "lines": "", "claim": ""}
  ],
  "uncertainties": []
}
```

Write the single game-level JSON object to the assigned OUTPUT_PATH using apply_patch. Every configuration_reports member must contain numeric case_index and difficulty. The game-level object must identify the assigned game and contain the complete requested configuration set. This one game report is the only permitted persistent output. Do not edit puzzle sources, definitions, historical audits, or another report. Do not inspect other first-pass reviews, existing classification/capability files, evaluation results, or previous conversations. The audit manifest is parent-only; the launcher gives you your exact assignment.

You may run read-only source/math probes when needed, with PYTHONDONTWRITEBYTECODE=1 and without modifying source. For a seed probe, include seeds, numerator/denominator, and reproducible probe code in pre_run.seed_probe. No browser or desktop access, external websites/services, nested agents, or empirical CUA evaluation. A source-review report may be unresolved; do not guess to finish.

files_read_completely and source_evidence must use repository-relative paths. Read each complete implementation file needed by the assigned mechanic, including shared helpers actually used. If a shared file is long, read it in successive ranges to EOF; keyword searches do not substitute for full reads. Record concrete code evidence for the exact difficulty and interaction. Do not make changes to obtain a preferred label.

Before finishing, load your JSON and check the game identity, exact requested configuration coverage, numeric IDs, label/clause consistency, and all cited file paths and line bounds. Shared evidence may be reused where it applies, but every configuration must remain independently understandable. A structural check is not evidence that the substantive classification is correct. Do not read other reviewers' reports or the audit manifest.

\end{Verbatim}

\section{Extended Related Work}
\label{app:related-work}

This appendix expands on the related work discussion in Section~\ref{sec:related-work}.

\paragraph{Benchmarks for interactive visual agents.}
Existing benchmarks for vision-language agents span web, desktop, mobile, game, and embodied environments. Web benchmarks such as WebArena and VisualWebArena provide self-hosted websites with realistic tasks \citep{zhou2024webarena,koh2024visualwebarena}, while desktop benchmarks such as OSWorld and Windows Agent Arena evaluate open-ended tasks across operating system applications \citep{xie2024osworldbenchmarkingmultimodalagents,bonatti2024windowsagentarena}. OSWorld 2.0 further extends this setting to 108 long-horizon workflows, where the best agent completes only 20.6\% of tasks \citep{yuan2026osworld2}. On mobile, AndroidWorld provides tasks across 20 Android apps \citep{rawles2024androidworld}, and Programming with Pixels and Gym-Anything extend computer-use evaluation to software engineering and to 200 professional software applications \citep{aggarwal2025programmingpixelscomputeruseagents,aggarwal2026gymanythingturnsoftwareagent}. Game benchmarks evaluate the same models on complete games, ranging from classic video games \citep{zhang2026videogamebenchvisionlanguagemodelscomplete} and reinforcement learning environments such as NetHack \citep{paglieri2025balrog} to platformers, puzzle games, and browser games \citep{hu2025lmgamebench,zheng2025vmage,ying2026aigamestorescalableopenended,ouyang2026gameworldstandardizedverifiableevaluation}. ARC-AGI-3 evaluates exploration and goal inference in turn-based abstract games, where humans solve all environments while frontier models solve less than 1\% \citep{arcprize2026arcagi3}. Embodied benchmarks similarly evaluate agents in simulated homes and physical environments \citep{savva2019habitatplatformembodiedai,Shridhar_2020_CVPR,srivastava2021behaviorbenchmarkeverydayhousehold,yang2025embodiedbenchcomprehensivebenchmarkingmultimodal}. However, since each task in these benchmarks requires many capabilities at once, low performance does not identify which capability limits the agent.

Several works attempt to attribute performance to individual capabilities. SmartPlay maps each of its six games to a subset of nine agent capabilities \citep{wu2023smartplay}, LVLM-Playground evaluates perception, question answering, rule following, and end-to-end play separately on the same board games \citep{wang2025lvlmplayground}, and Embodied Agent Interface decomposes embodied decision making into separate modules, arguing that final success rates cannot pinpoint which ability is missing \citep{li2024embodiedagentinterface}. VS-Bench similarly reports perception, strategic reasoning, and decision making separately \citep{xu2026vsbench}. Other benchmarks vary conditions of the same task, such as difficulty tiers \citep{davydova2025osuniverse,zheng2025mcu}, initial states \citep{zhao2025worldgui}, or interface changes between actions \citep{liu2026dynamicguibench}. However, these works either operate through text or turn-based interfaces, or vary a single condition without controlling the capabilities a task requires. In contrast, \bench{} annotates the capabilities required by each configuration, and varies difficulty, interaction interface, and real-time setting on the same minigame.

\paragraph{Capability-specific evaluation of VLMs.}
A large body of work evaluates individual capabilities of VLMs through question answering. For 3D spatial understanding, VSI-Bench, MindCube, 3DSRBench, and OmniSpatial evaluate spatial relations, viewpoint changes, and mental models of scenes \citep{Yang_2025_CVPR,wang2025mindcube,ma2024threedsrbench,jia2025omnispatial}, while BLINK shows that perception tasks that humans solve easily remain difficult for multimodal models \citep{fu2024blink}. For temporal reasoning, TempCompass, TOMATO, and Vinoground evaluate speed, direction, and event order in videos \citep{liu-etal-2024-tempcompass,shangguan2024tomato,zhang2024vinoground}, with TOMATO showing that many questions in earlier video benchmarks can be answered from a single frame. For GUI grounding, SeeClick and ScreenSpot-Pro evaluate whether models can locate screen elements from instructions \citep{cheng-etal-2024-seeclick,li2025screenspot}. Finally, PuzzleVQA, AlgoPuzzleVQA, and VisualPuzzles evaluate abstract and algorithmic reasoning over images \citep{chia2024puzzlevqa,ghosal2024language,song2025visualpuzzles}, and MMMU and MMMU-Pro evaluate multidisciplinary reasoning over images and text \citep{Yue_2024_CVPR,yue-etal-2025-mmmu}. However, in all of these benchmarks the model answers a question about a static input, without having to act on its understanding. Closer to our work, iVISPAR evaluates the spatial reasoning of VLM agents on a sliding-tile puzzle presented in 3D, 2D, and text \citep{mayer2025ivispar}, and VisGym provides 17 interactive environments with configurable difficulty and input representation \citep{wang2026visgym}. \bench{} complements these works by covering temporal reasoning, exploration and interface understanding, and real-time interaction, and by annotating the capabilities required by each configuration.

\paragraph{Real-time interaction.}
Acting in dynamic environments requires accounting for the delay between observing the environment and executing an action. In robotics, real-time action chunking and asynchronous inference execute previously planned actions while the next chunk is computed \citep{black2025realtimeexecutionactionchunking,shukor2025smolvla}, and VLASH conditions the policy on the state predicted at execution time \citep{tang2025vlash}. Reinforcement learning has similarly studied decision making under action and observation delays \citep{ramstedt2019realtime,derman2021acting}. For computer-use agents, OSWorld-Human measures end-to-end latency and finds that agents take 2.7 to 4.3 times more steps than necessary \citep{abhyankar2026osworldhumanbenchmarkingefficiencycomputeruse}. Game benchmarks have taken different approaches to latency. GameWorld pauses the game during inference and reports a separate real-time study with two models \citep{ouyang2026gameworldstandardizedverifiableevaluation}, VideoGameBench runs games in real time and additionally provides a Lite version that pauses the emulator while the agent is thinking \citep{zhang2026videogamebenchvisionlanguagemodelscomplete}, and Orak compares paused and real-time play on StarCraft~II \citep{park2025orak}. Real-Time Reasoning Gym advances text-based games after a fixed number of generated tokens, comparing reactive and planning agents under increasing time pressure \citep{wen2026realtime}. Our work complements these efforts by evaluating the same minigame and seed in the live, slowed, and paused settings, and by pairing each real-time minigame with a non-real-time counterpart.

\paragraph{Automated environment construction.}
A growing body of work uses coding agents and LLMs to construct environments for agents. Gym-Anything uses a creation agent and an independent audit agent to convert 200 software applications into computer-use environments \citep{aggarwal2026gymanythingturnsoftwareagent}. CUA-Gym separates an agent that constructs environment states from one that writes reward functions \citep{wang2026cuagym}, and GUI-GENESIS converts real applications into lightweight web environments with executable rewards \citep{cao2026guigenesis}. AutoEnv generates 36 text-based environments with 358 levels \citep{zhang2025autoenv}, while AgentGen and OMNI-EPIC generate environments together with tasks of varying difficulty \citep{hu2024agentgen,faldor2024omniepic}. Similar pipelines construct software engineering tasks from code repositories \citep{pan2025swegym,yang2025swesmith} and 3D scenes for embodied agents \citep{yang2024holodeck}. Procedural generation further provides task instances with adjustable complexity, including Procgen games \citep{cobbe2020leveraging}, MiniHack and BabyAI \citep{samvelyan2021minihack,chevalierboisvert2019babyai}, and verifiable reasoning problems in Reasoning Gym and Enigmata \citep{stojanovski2025reasoning,chen2025enigmata}. However, these approaches construct software, text, or game environments, and do not produce interactive visual minigames whose difficulty and interaction interface can be varied while preserving the underlying task. We extend the creator-auditor approach to such minigames, where each minigame has a non-real-time counterpart, five difficulty levels, two interaction modes, and an oracle solution.

\paragraph{CAPTCHAs and visual puzzles.}
Many of the seed minigames in \bench{} are derived from CAPTCHAs and puzzle games. CAPTCHA research has studied whether automated solvers can pass challenges designed to distinguish humans from bots. Recent works solve reasoning and visual CAPTCHAs with LLMs and VLMs \citep{deng2024oedipus,teoh2025captchas}, analyze which types of challenges remain difficult for these models \citep{wang2025cognition,wu2025mcabench}, and train GUI agents on interactive CAPTCHAs \citep{chen2026captcha}. Open CaptchaWorld evaluates web agents on 20 types of interactive CAPTCHAs, where the best agent achieves 40.0\% compared to 93.3\% for humans \citep{luo2025opencaptchaworld}. Beyond CAPTCHAs, KORGym, DiG-bench, and VGRP-Bench evaluate multi-turn reasoning in text, game, and grid puzzles \citep{shi2025korgym,battleday2026dig,ren2025vgrpbench}. However, these works primarily target security or reasoning in isolation. In contrast, \bench{} uses such challenges as seed minigames, and adds controlled capability requirements, difficulty levels, interaction interfaces, and real-time settings.

\end{document}